\documentclass[authoryear,3p,final,times,10pt]{elsarticle}
\usepackage{amssymb}
\usepackage[T1]{fontenc}
\usepackage[utf8]{inputenc}

\usepackage[linesnumbered, ruled, vlined]{algorithm2e} 
\usepackage{amsmath, amsfonts} % For math environments

\usepackage[noadjust]{cite}

\usepackage{apalike}
\usepackage{float}
\usepackage{makecell}
\usepackage{multirow}
\usepackage{adjustbox}
\usepackage{lscape}
\usepackage{float}
\usepackage[labelfont=bf]{caption}
\usepackage{subcaption}
\usepackage{amsmath}
\usepackage{amsfonts}
\usepackage{sidenotes}
\usepackage[hidelinks]{hyperref}
\hypersetup{
  colorlinks   = true, %Colours links instead of ugly boxes
  urlcolor     = blue, %Colour for external hyperlinks
  linkcolor    = blue, %Colour of internal links
  citecolor   = blue %Colour of citations
}
\usepackage{pifont}
\usepackage{graphicx}
\usepackage{hyperref}
\usepackage{xcolor}
\usepackage{longtable}
\usepackage{color}
\usepackage{xcolor,soul,framed} %,caption

\colorlet{shadecolor}{yellow}

\begin{document}

\begin{frontmatter}

%% Title, authors and addresses

%% use the tnoteref command within \title for footnotes;
%% use the tnotetext command for theassociated footnote;
%% use the fnref command within \author or \affiliation for footnotes;
%% use the fntext command for theassociated footnote;
%% use the corref command within \author for corresponding author footnotes;
%% use the cortext command for theassociated footnote;
%% use the ead command for the email address,
%% and the form \ead[url] for the home page:
%% \title{Title\tnoteref{label1}}
%% \tnotetext[label1]{}
%% \author{Name\corref{cor1}\fnref{label2}}
%% \ead{email address}
%% \ead[url]{home page}
%% \fntext[label2]{}
%% \cortext[cor1]{}
%% \affiliation{organization={},
%%            addressline={}, 
%%            city={},
%%            postcode={}, 
%%            state={},
%%            country={}}
%% \fntext[label3]{}

% \title{DGNN-YOLO: Interpretable Dynamic Graph Neural Networks with YOLO11 for Detecting and Tracking Small Occluded Objects in Urban Traffic}
\title{Interpretable Dynamic Graph Neural Networks for Small Occluded Object Detection and Tracking}

% \iffalse
\author[1]{Shahriar Soudeep\corref{contrib}}
\ead{20-43823-2@student.aiub.edu}

\author[2,3]{Md Abrar Jahin\corref{contrib}}
\ead{abrar.jahin.2652@gmail.com}

\author[1]{and M. F. Mridha\corref{corauthor}}
\ead{firoz.mridha@aiub.edu}

%\author[3]{and Nilanjan Dey}
%\ead{nilanjan.dey@tint.edu.in}

\address[1]{Department of Computer Science, American International University-Bangladesh, Dhaka, 1229, Bangladesh}
\address[2]{Department of Industrial Engineering and Management, Khulna University of
Engineering and Technology (KUET), Khulna 9203, Bangladesh}
\address[3]{Physics and Biology Unit, Okinawa Institute of Science and Technology Graduate University (OIST), Okinawa, 904-0412, Japan}

%\address[3]{Department of Computer Science \& Engineering, Techno International New Town, New Town, Kolkata, 700156, India}

\cortext[corauthor]{Corresponding author}
\cortext[contrib]{Authors contributed equally}
% \fi

\begin{abstract}
The detection and tracking of small, occluded objects—such as pedestrians, cyclists, and motorbikes—pose significant challenges for traffic surveillance systems because of their erratic movement, frequent occlusion, and poor visibility in dynamic urban environments. Traditional methods like YOLO11, while proficient in spatial feature extraction for precise detection, often struggle with these small and dynamically moving objects, particularly in handling real-time data updates and resource efficiency. This paper introduces DGNN-YOLO, a novel framework that integrates dynamic graph neural networks (DGNNs) with YOLO11 to address these limitations. Unlike standard GNNs, DGNNs are chosen for their superior ability to dynamically update graph structures in real-time, which enables adaptive and robust tracking of objects in highly variable urban traffic scenarios. This framework constructs and regularly updates its graph representations, capturing objects as nodes and their interactions as edges, thus effectively responding to rapidly changing conditions. Additionally, DGNN-YOLO incorporates Grad-CAM, Grad-CAM++, and Eigen-CAM visualization techniques to enhance interpretability and foster trust, offering insights into the model’s decision-making process. Extensive experiments validate the framework’s performance, achieving a precision of 0.8382, a recall of 0.6875, and mAP@0.5:0.95 of 0.6476, significantly outperforming existing methods. This study offers a scalable and interpretable solution for real-time traffic surveillance and significantly advances intelligent transportation systems' capabilities by addressing the critical challenge of detecting and tracking small, occluded objects.
\end{abstract}

\iffalse

\begin{highlights}
\item Research highlight 1
\item Research highlight 2
\end{highlights}
\fi


\end{frontmatter}

%% \linenumbers
\section{Introduction}
The rapid development of intelligent transportation systems (ITS) has significantly transformed urban traffic management and mobility. ITS leverages advanced technologies designed to improve road safety, reduce congestion, and enable real-time decision-making in urban environments~\citep{Furda2011}. One key component of ITS is the detection and tracking of small objects, such as pedestrians, cyclists, and motorbikes, which are essential for traffic monitoring, accident prevention, and autonomous vehicle operations. However, detecting and tracking these small objects is challenging due to factors like occlusion, low resolution, fluctuating lighting conditions, and high object density. These issues often render traditional detection methods insufficient in dynamic, real-world settings~\citep{sun_rsod_2022}.

In the context of sustainable urban mobility, enhancing traffic surveillance systems is vital for safety and efficiency~\citep{elassy_intelligent_2024}. Traditional object detection and tracking methods, which relied on handcrafted features and traditional machine learning, often struggled in dynamic traffic scenarios due to limited generalizability~\citep{fiaz_handcrafted_2019}. The advent of deep learning (DL), particularly convolutional neural networks (CNNs), improved detection performance by learning complex spatial features directly from data~\citep{yadav2024}. However, CNNs face limitations in capturing temporal relationships essential for object tracking across video frames~\citep{chu2017online}. To overcome these shortcomings, graph neural networks (GNNs) have emerged as a promising approach for dynamically modeling spatial-temporal interactions, enabling robust and real-time tracking~\citep{shao2022decoupled}. Recent efforts to integrate YOLO and GNNs have shown potential, but their opaque decision-making processes pose challenges for interpretability and trust in real-world applications~\citep{degas_survey_2022}.

\begin{figure}[!ht]
    \centering
    \includegraphics[width=0.7\linewidth]{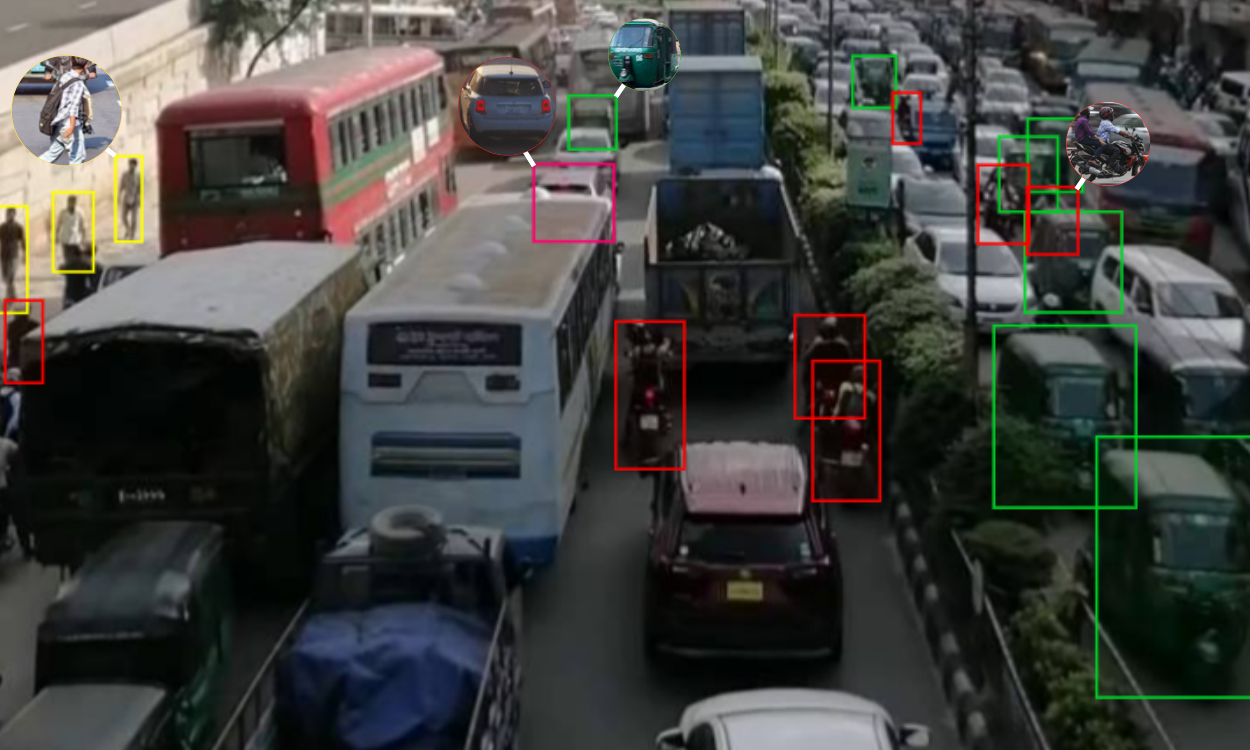}
    \caption{A snapshot of an urban traffic environment illustrating challenges in detecting and tracking small, occluded objects. Green bounding boxes highlight rickshaws and small vehicles; red boxes mark motorcycles; yellow boxes indicate pedestrians; pink boxes correspond to larger vehicles, such as trucks. Occlusion, congestion, and unpredictable movements of small objects complicate reliable detection and tracking, emphasizing the need for advanced surveillance systems to improve safety and traffic flow.}
    \label{fig:introduction_fig}
\end{figure}

In urban traffic environments, as shown in Figure~\ref{fig:introduction_fig}, small objects, particularly partially or fully occluded ones, present significant challenges for detection and tracking. These objects often fail to adhere to established traffic guidelines, unpredictably changing lanes. Such erratic behavior disrupts the flow of larger vehicles, necessitating frequent braking. This cascading effect forces other vehicles within the same lane to brake, leading to congestion. The problem is further exacerbated when these objects encroach upon pedestrian walkways. This forces pedestrians to step onto the road, creating safety risks and further compounding traffic congestion. Addressing this issue requires robust systems capable of reliably detecting and tracking small, occluded objects. However, achieving this is particularly challenging due to their dynamic behavior and weak visibility cues. Our proposed model addresses this gap by providing effective detection and tracking capabilities for such objects. It is designed to ensure comprehensive monitoring and adherence to traffic guidelines, enabling the implementation of timely corrective measures. This approach improves traffic safety and mitigates congestion caused by unpredictable behaviors of small, occluded objects.

This paper introduces the DGNN-YOLO framework, which integrates YOLO11 for small-object detection with Dynamic Graph Neural Networks (DGNN) for robust tracking. YOLO11, the latest iteration of the "You Only Look Once" (YOLO) family, is optimized for real-time applications through advanced spatial feature extraction and fine-grained anchor box design, enabling superior detection of small and occluded objects in dynamic traffic scenarios~\citep{alif_yolov11_2024}. The DGNN component leverages dynamic graph structures, where nodes represent detected objects and edges model spatiotemporal relationships based on motion trajectories and interaction patterns. These graphs are updated in real-time, ensuring adaptive tracking across video frames even in highly congested and unpredictable environments~\citep{yuan_temporal_2017}. To enhance interpretability, we incorporate explainable AI (XAI) techniques, including Grad-CAM, Grad-CAM++, and Eigen-CAM, to visualize critical spatial and temporal features influencing the model’s predictions. These visual explanations provide insights into both detection and tracking processes, increasing the transparency and trustworthiness of the framework~\citep{mankodiya_od-xai_2022}. The DGNN-YOLO framework is rigorously evaluated on the \textit{i2 Object Detection Dataset}, a benchmark designed for traffic surveillance that includes challenges such as heavy occlusion, rapid object movements, and diverse lighting conditions. This evaluation demonstrates the framework’s ability to achieve accurate and interpretable predictions, establishing its reliability and scalability for real-world traffic monitoring applications.

The main contributions of this study are summarized as follows:
\begin{enumerate}
    \item We propose a novel DGNN-YOLO framework that integrates YOLO11 and DGNN for the real-time detection and tracking of small objects, addressing challenges such as occlusion and motion blur.
    \item A dynamic graph construction and update mechanism is introduced to effectively model spatial-temporal relationships, improving tracking accuracy in complex traffic environments.
    \item The proposed framework was comprehensively evaluated using the \textit{i2 Object Detection Dataset}, showing significant performance improvements over existing methods in detecting and tracking small objects.
    \item We incorporate XAI techniques (Grad-CAM, Grad-CAM++, and Eigen-CAM) to enhance the interpretability of the detection and tracking process.
\end{enumerate}

The rest of this paper is organized as follows: Section~\ref{sec:related_work} discusses related work on small-object detection and tracking. Section~\ref{sec:methodology} elaborates on the methodology and details the integration of YOLO11 and the DGNN within the DGNN-YOLO framework and the model-agnostic XAI techniques implemented in our research. Section~\ref{sec:experiments} presents the experimental setup and results, including interpretability analysis. Finally, Section~\ref{sec:conclusion} concludes the study with key findings and potential future research directions.

\section{Related Work}
\label{sec:related_work}
Small object detection and tracking fields have witnessed significant advancements driven by challenges such as occlusion, low resolution, and dynamic environments. Accurate detection and tracking of small objects are critical for applications such as traffic surveillance and autonomous systems; however, traditional methods often struggle with generalization in complex real-world scenarios. Recent progress in DL has led to specialized techniques for small object detection and dynamic interaction modeling, while the adoption of GNNs has provided an effective approach for capturing spatial-temporal relationships in tracking tasks. This section reviews the key developments in these areas, which form the foundation of the proposed framework.

\subsection{Small Object Detection}
Owing to occlusion, low resolution, and scale variation, small-object detection has been a persistent challenge, particularly in traffic surveillance. Early methods relied on handcrafted features, such as histogram of oriented gradients (HOG) and scale-invariant feature transform (SIFT), combined with classifiers like support vector machines (SVMs)~\citep{dalal2005histograms}. While effective in controlled environments, these approaches often failed in complex real-world scenarios with variable lighting and motion~\citep{viola2001rapid}. While YOLO-Anti demonstrates significant improvements in handling congested scenes, its ability to detect small objects may still be limited by the inherent challenges of balancing foreground and background features, as well as the potential loss of fine-grained details during feature aggregation across different levels~\citep{WANG2022108814}. Although BiFPN-YOLO shows enhanced performance through improved feature fusion with BiFPN, small object detection remains a challenge due to the trade-off between increasing model complexity and maintaining computational efficiency, which can limit the granularity of feature extraction for tiny objects~\citep{DOHERTY2025111209}. Recent studies in camouflaged object detection have explored feature fusion, self-similarity constraints, and edge enhancement techniques to improve the localization of concealed objects, yet challenges persist in distinguishing subtle foreground-background differences, especially for small-scale objects~\citep{yang2025}. Existing object detection methods in foggy conditions rely on preprocessing and visual features, but Dehazing \& Reasoning YOLO (DR-YOLO) improves detection by integrating an atmospheric scattering model and co-occurrence relation graph as prior knowledge, improving accuracy without increasing inference time~\citep{zhong2024110756}. STDnet-ST improves small object detection in videos by leveraging a spatio-temporal convolutional network that links high-confidence detections across time, forming tubelets to improve accuracy while filtering unprofitable object associations~\citep{BOSQUET2021107929}.

The emergence of DL has revolutionized object detection. Ren et al. introduced Faster R-CNN, a two-stage detector that integrates region proposal networks with convolutional layers to enhance accuracy and speed~\citep{ren_faster_2017}. Redmon et al. developed YOLO, a single-stage framework that unifies detection and classification tasks, significantly improving real-time detection~\citep{redmon2016you}. However, earlier YOLO versions struggled with small-object detection due to their reliance on coarse-grained feature maps. Subsequent iterations, such as YOLOv3 and YOLOv4, addressed these limitations by incorporating multi-scale feature fusion and improved anchor mechanisms~\citep{}. Building on these advancements, YOLO11 introduces attention mechanisms and fine-grained feature extraction, significantly enhancing detection capabilities in dynamic environments. These features make YOLO11 an ideal candidate for urban traffic surveillance systems, particularly for detecting small objects like pedestrians, cyclists, motorbikes, and various vehicles, including auto-rickshaws and covered vans.

In addition to YOLO, other frameworks have contributed to small-object detection. RetinaNet employs focal loss to address class imbalance~\citep{ross2017focal}, while Single Shot MultiBox Detector (SSD) utilizes multi-scale feature maps for accurate localization~\citep{liu2016ssd}. Despite their strengths, achieving real-time detection under challenging conditions such as low light, rain, and heavy traffic remains difficult~\citep{chen_object_2019}. Emerging transformer-based architectures, such as detection transformers (DETRs), show promise in modeling global relationships for small-object detection, although their high computational requirements limit practical deployment~\citep{carion2020end}.

\subsection{Object Tracking}
Tracking small objects in traffic surveillance involves maintaining their identities across consecutive frames under challenging conditions, such as occlusion, abrupt motion, and dense traffic environments. Early methods, such as those reviewed by Marvasti-Zadeh \emph{et al.}, relied heavily on handcrafted features and motion models. While these methods laid foundational work, they often struggled to adapt to dynamic and complex scenarios due to their inflexibility and low generalization capability~\citep{marvasti-zadeh_deep_2022}. Recent advancements in DL have significantly improved tracking performance. Transformer-based approaches, for instance, have been particularly impactful. Chen \emph{et al.} introduced a transformer tracking framework that models long-range dependencies and contextual relationships across frames, achieving state-of-the-art performance~\citep{chen2021transformer}. However, these models typically require substantial computational resources, limiting their applicability in real-time or resource-constrained environments. Similarly, Blatter \emph{et al.} proposed a lightweight transformer architecture that balances efficiency and accuracy, making it more suitable for real-time applications in resource-constrained environments~\citep{blatter_efficient_2023}. While this approach improves efficiency, it may still fall short in handling the high variability and rapid dynamics typical of dense urban traffic scenarios.

Hybrid frameworks like SORT and DeepSORT combine detection and tracking to handle multi-object tracking (MOT) challenges. Wang \emph{et al.} developed a real-time MOT framework that integrates motion prediction and appearance modeling to improve scalability and robustness in dynamic scenarios~\citep{wang2020towards}. Although effective, these methods can struggle with persistent occlusions and overlapping objects, areas where our research introduces novel methodologies to enhance detection accuracy and tracking consistency. Modern transformer-based trackers like SwinTrack, introduced by Lin \emph{et al.}, utilize hierarchical feature representation and cross-scale attention mechanisms to enhance tracking robustness, especially in crowded environments~\citep{lin_swintrack_2022}. However, these sophisticated models can be overly reliant on high-quality data and may not perform as well in low-visibility conditions, a gap our current study addresses by implementing adaptive lighting compensation techniques.

Comprehensive reviews by Aziz \emph{et al.} provide an overview of DL-based object tracking~\citep{aziz_exploring_2020}. They highlight the progress in the field and discuss challenges such as scalability and efficiency. Yet, these reviews often overlook the need for integrating multi-modal data to improve the contextual understanding of scenes, a limitation our study seeks to overcome by proposing a generalized tracking framework that leverages both visual and sensor data for enhanced object identification and tracking.

\subsection{GNNs in Object Tracking}
GNNs have emerged as a robust framework for modeling spatial-temporal relationships in object-tracking tasks. By representing detected objects as nodes and their interactions as edges, GNNs effectively capture dependencies across consecutive frames, making them particularly suitable for dynamic and complex environments. Jiang \emph{et al.} demonstrated the potential of GNNs by integrating object detection and multi-object tracking into a unified framework, achieving enhanced tracking accuracy and robustness. Despite their success, these frameworks often face computational challenges, particularly in real-time applications, due to their high processing demands~\citep{jiang2019graph}. Our research directly addresses these computational inefficiencies by introducing a novel graph pruning algorithm that reduces the complexity of the GNN computations without compromising the accuracy of the tracking. This innovation allows our model to operate efficiently in real-time scenarios, overcoming one of the primary limitations faced by the approaches developed by Zhang \emph{et al.}.

In multi-object tracking, the work by Weng \emph{et al.} on GNN3DMOT utilizes multi-feature learning to improve tracking performance under challenging conditions. However, their approach can be computationally intensive and less effective in densely populated scenes~\citep{weng2020gnn3dmot}. Building upon this, B{\"u}chner and Valada have further refined GNN applications in 3D tracking with their cross-edge modality attention mechanism, which enhances precision but still struggles with rapid scene changes~\citep{buchner20223d}. Our contribution extends these methodologies by incorporating an adaptive feature selection mechanism that dynamically adjusts the features used based on the scene complexity, significantly improving processing speed and accuracy in high-density scenarios. Moreover, adaptive graph construction techniques like those in Zhang \emph{et al.}'s SCGTracker and Ma \emph{et al.}'s Deep Association have improved handling occlusions and complex object interactions. However, they are susceptible to errors in environments with poor sensor quality or rapid object movements~\citep{zhang2024scgtracker}. Our system enhances these models by integrating a real-time environmental feedback module that adjusts the graph parameters dynamically, ensuring robust tracking even under fluctuating environmental conditions.

Lastly, the advancements in data association methods by Lee \emph{et al.} have showcased improved online multi-object tracking performance. Nonetheless, these systems require frequent updates and recalibrations to remain effective across different tracking conditions~\citep{lee2021graph}. We tackle this challenge by developing a self-learning graph convolutional network that continuously evolves based on incoming data, drastically reducing the need for manual recalibration and enhancing the system’s adaptability across varied conditions.

By addressing these challenges, our research significantly advances the practical deployment and operational efficiency of GNN-based object tracking systems, providing robust solutions to the scalability, accuracy, and adaptability issues that have previously hindered broader applications.

\section{Methodology}  
\label{sec:methodology}  
The proposed system for detecting and tracking small objects in traffic surveillance comprises four interconnected components designed to address challenges such as occlusion, low resolution, and motion blur. DGNN-YOLO provides a high-level integration of detection and tracking mechanisms to handle real-time processing demands. The YOLO11 detection mechanism is a state-of-the-art module for identifying small objects in complex traffic environments. Dynamic graph construction represents detected objects and their spatial-temporal relationships using dynamic graph structures, enabling the adaptive modeling of interactions. Finally, the DGNN-based tracking module employs DGNNs to refine object associations across frames, ensuring robust and accurate tracking. These components enable efficient video data processing and reliable performance under diverse and challenging traffic conditions.

\subsection{Notations}
This section introduces the symbols and background information used in the study. Table~\ref{tab:notations} provides a detailed list of the commonly used symbols and their definitions, offering a clear understanding of the mathematical and structural components of the proposed DGNN-YOLO framework. Additionally, it includes the symbols and mathematical notations relevant to the XAI techniques—Grad-CAM, Grad-CAM++, and Eigen-CAM used to interpret the model's predictions. These notations form the foundation for describing the integration of YOLO11, DGNN, and XAI techniques and their application to small-object detection, tracking, and explainability in traffic surveillance.

\begin{table}[!ht]
\centering
\footnotesize
\caption{Frequently used notations and their definitions for the DGNN-YOLO framework for small object detection and tracking.}
\begin{tabular}{cl}
\hline
\textbf{Symbol} & \textbf{Definition} \\ \hline
\( I_t \)       & The input video frame at time \( t \) \\ 
\( D_t \)       & The set of objects detected in frame \( I_t \) by YOLO11 \\ 
\( B_i \)       & The bounding box of object \( i \) in \( D_t \), defined as \([x, y, w, h]\)  \\ 
\( \mathcal{C}_i \) & The confidence score of object \( i \) in \( D_t \)  \\ 
\( L_i \)       & The class label of object \( i \) in \( D_t \), such as ``CNG" or ``motorbike"  \\ 
\( G_t \)       & The dynamic graph at time \( t \), constructed for tracking  \\ 
\( N_t \)       & The set of nodes in \( G_t \), each representing a detected object  \\ 
\( E_t \)       & The set of edges in \( G_t \), representing interactions between nodes  \\ 
\( x_i \)       & The feature vector of node \( i \) in \( G_t \)  \\ 
\( e_{ij} \)    & The edge weight between nodes \( i \) and \( j \), capturing interaction strength  \\ 
\( A_t \)       & The adjacency matrix of \( G_t \), encoding edge connections between nodes  \\ 
\( F_s \)       & The spatial features extracted by YOLO11 for each detected object  \\ 
\( F_t \)       & The temporal features computed by DGNN for tracking objects across frames  \\ 
\( \mathcal{L}_{\text{det}} \) & The loss function for YOLO11 detection  \\ 
\( \mathcal{L}_{\text{track}} \) & The loss function for DGNN-based tracking  \\ 
\( \tau \)      & The threshold for object detection confidence  \\ 
\( T \)         & The total number of frames processed in the video sequence  \\ 
\( R \)         & The region of interest (ROI) for detecting and tracking objects  \\ 
\( v_{i,t} \)   & The velocity of object \( i \) at time \( t \)  \\ 
\( \Delta t \)  & The time difference between consecutive frames  \\ 
\( P_t \)       & The object position matrix at time \( t \), describing the spatial locations of all objects  \\ 
\( \mathcal{M} \) & The mapping function linking detections to graph nodes  \\ 
\( \mathcal{E} \) & The error term quantifying tracking inconsistency  \\ 
\( L_{\text{Grad-CAM}}^c \) & Localization map for class \( c \) generated by Grad-CAM \\ 
\( \alpha_k^c \) & Weight of the \( k \)-th feature map for class \( c \) in Grad-CAM/Grad-CAM++ \\ 
\( A^k \) & Activation feature map of the \( k \)-th channel \\ 
\( y^c \) & Output score for the target class \( c \) \\ 
\( \text{EigenVec}(A) \) & Principal eigenvector of the activation matrix \( A \) in Eigen-CAM \\ \hline
\end{tabular}
\label{tab:notations}
\end{table}

\subsubsection{Input Video and Object Detection}

The input video sequence is denoted as \( \{ I_t \}_{t=1}^T \), where \( T \) is the total number of frames. Each frame \( I_t \) undergoes processing by the YOLOv11 model for object detection, resulting in a set of detected objects \( D_t = \{ (B_i, C_i, L_i) \}_{i=1}^{N_t} \), where \( N_t \) is the number of objects detected in frame \( I_t \). The bounding box for each object \( i \) is represented by \( B_i = [x_i, y_i, w_i, h_i] \), where \( (x_i, y_i) \) are the coordinates of the center, and \( w_i, h_i \) are the width and height. The confidence score \( C_i \) reflects the model's certainty in the detection of the object, and the class label \( L_i \) identifies the object category (e.g., "auto-rickshaw," "bus").

Additionally, YOLOv11 extracts spatial features \( F_s \) for all detected objects. These features encapsulate the geometric and appearance-based information essential for the subsequent tracking phase. These spatial features are fed into the graph construction process, which models the relationships between objects across video frames, enabling accurate tracking over time.

\subsubsection{Dynamic Graph Construction}

For each frame \( t \), a dynamic graph \( G_t = (N_t, E_t) \) is constructed to represent the objects and their interactions. The set of nodes, \( N_t = \{ n_i \}_{i=1}^{N_t} \), corresponds to the detected objects, each node \( n_i \) represented by a feature vector \( x_i = [F_s, F_t] \), which combines spatial features from YOLOv11 and temporal features computed by the DGNN. These features model the evolution of the object's positions and interactions over time, capturing both spatial and temporal dynamics.

The edges \( E_t \) model the relationships between objects. The adjacency matrix \( A_t \), which encodes the strength of these relationships, is updated dynamically to reflect changes in object positions, velocities, and appearance. This dynamic update mechanism allows the model to adapt to real-time changes in the scene, ensuring accurate tracking even in dense and unpredictable traffic conditions.

\subsubsection{Node and Edge Features}

Each node \( n_i \) in the graph is represented by a feature vector \( x_i = [F_s, F_t] \), where \( F_s \) captures the spatial features extracted by YOLOv11 and \( F_t \) encodes temporal information derived from the DGNN. These temporal features represent the movement and interaction of objects across consecutive frames, providing the model with a dynamic understanding of the scene.

The edges between nodes are weighted based on several factors. The proximity between two objects is captured by the Euclidean distance \( d_{ij} = \| (x_i, y_i) - (x_j, y_j) \|_2 \), representing the spatial closeness. Velocity similarity is computed as \( \Delta v_{ij} = \| v_i(t) - v_j(t) \|_2 \), where \( v_i(t) \) and \( v_j(t) \) are the velocities of objects \( i \) and \( j \), respectively. Additionally, appearance similarity is determined through the cosine similarity of the feature embeddings \( f_i \) and \( f_j \), derived from the bounding box features. These factors contribute to the weighted adjacency matrix \( A_t \), which encodes the strength of the object interactions, ensuring robust associations across frames.

\subsubsection{Spatial-Temporal Interaction}

The DGNN-YOLO model leverages spatial-temporal interactions to effectively track objects over time. The spatial features \( F_s \), extracted from the bounding boxes by YOLOv11, enable precise localization and identification of objects, facilitating accurate detection. On the other hand, the temporal features \( F_t \) computed by the DGNN capture the evolution of object positions and interactions over time, modeling their motion dynamics across frames.

This dual integration of spatial and temporal features allows the graph structure to evolve dynamically. As objects move, the edges are updated to reflect changes in object relationships, ensuring that the model can track objects robustly even in rapidly changing and cluttered traffic environments.

\subsubsection{Loss Functions}

The DGNN-YOLO framework optimizes two primary loss functions to ensure accurate detection and robust tracking. The detection loss \( L_{\text{det}} \) evaluates the accuracy of object detection. It is composed of a bounding box loss (typically \( L_1 \)-loss) and a classification loss (cross-entropy loss) and is given by:
\begin{equation}
L_{\text{det}} = L_{\text{bbox}} + L_{\text{cls}}
\end{equation}

The tracking loss \( L_{\text{track}} \) ensures consistency in object identities across frames by minimizing inconsistencies in temporal associations and maintaining node and edge feature consistency. The tracking loss is computed as:
\begin{equation}
L_{\text{track}} = \sum_{(i,j) \in E_t} \| x_i - x_j \|_2^2 + \lambda_{\text{track}} \sum_i \| f_i^{t+1} - f_i^t \|_2^2
\end{equation}
where \( \lambda_{\text{track}} \) is a regularization term. The total loss is the weighted sum of the detection and tracking losses:
\begin{equation}
L_{\text{total}} = \lambda_{\text{det}} L_{\text{det}} + \lambda_{\text{track}} L_{\text{track}}
\end{equation}
where \( \lambda_{\text{det}} \) and \( \lambda_{\text{track}} \) are hyperparameters that balance the contributions of the detection and tracking tasks.

\subsubsection{Real-Time Processing and Regions of Interest}

The DGNN-YOLO system processes frames sequentially, focusing on a predefined Region of Interest (ROI) \( R \) to enhance computational efficiency. By prioritizing areas with high traffic activity, the model ensures that computational resources are used where they are most needed. This optimization reduces the processing load, allowing the system to operate in real-time even in environments with high object density. The ROI is updated dynamically as objects enter and exit the scene, ensuring that the system remains efficient while maintaining high accuracy in detection and tracking. This real-time adaptability allows DGNN-YOLO to operate effectively in dynamic and complex traffic conditions.

\begin{figure}[!ht]
    \centering
    \includegraphics[width=0.9\linewidth]{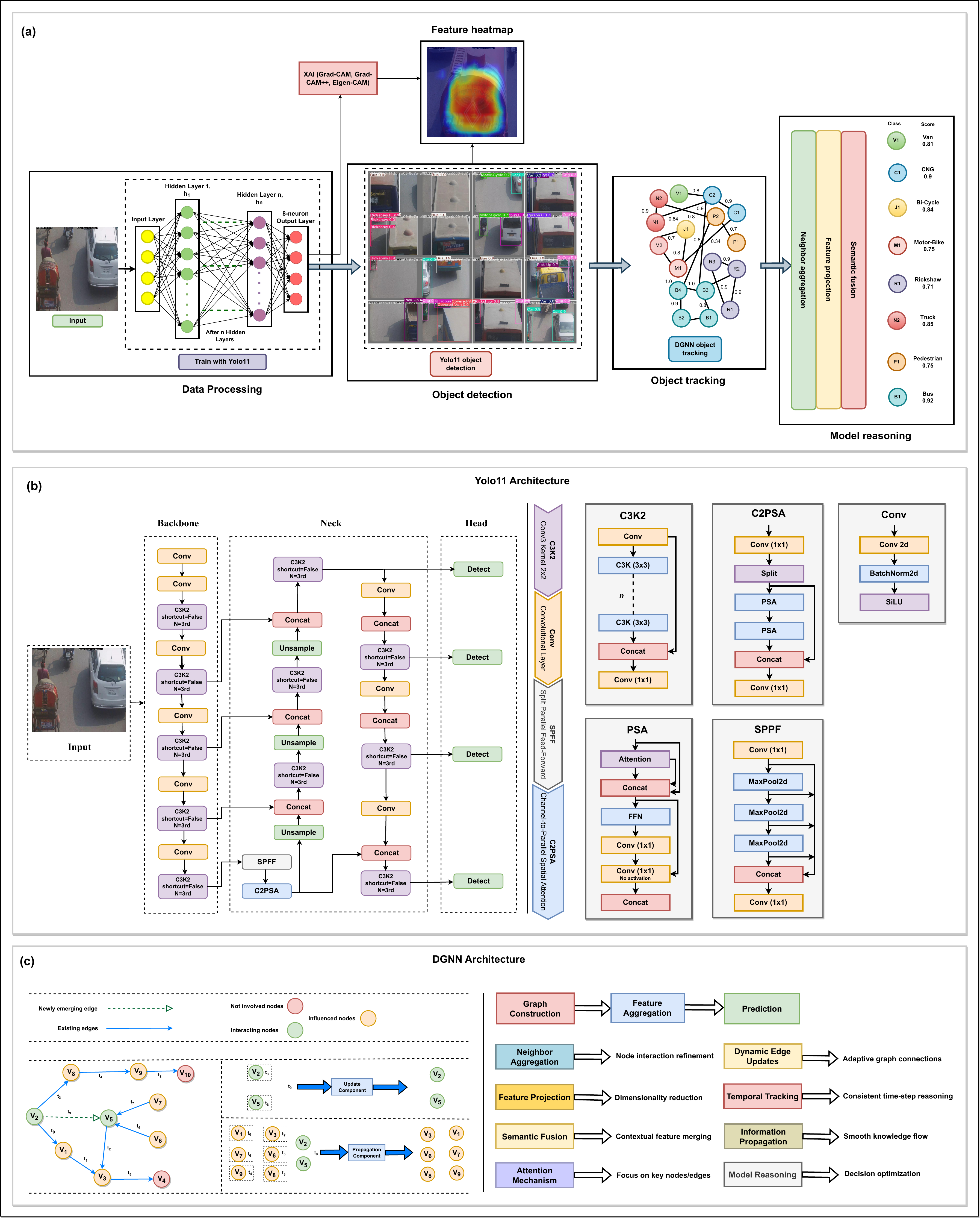}
    \caption{(a) Overview of the proposed DGNN-YOLO framework for small object detection and tracking in traffic videos, (b) YOLO11 architecture for small object detection, and (c) DGNN architecture for object tracking.}
    \label{fig:whole_process}
\end{figure}

\subsection{Overview of the Proposed Framework}
The DGNN-YOLO framework, shown in Figure~\ref{fig:whole_process}, integrates the YOLOv11 object detection model and the Dynamic Graph Neural Network (DGNN) for small object detection and tracking. The process begins by passing video frames through the YOLOv11 model, which detects objects, draws bounding boxes, and assigns confidence scores and class labels to the detected items. These detection outputs, including bounding box coordinates and class labels, are then utilized to construct a dynamic graph. Each detected object is represented as a node in the graph, and the relationships between these objects are captured as edges that model spatial-temporal interactions, such as proximity and velocity similarity. The graph is updated as the video progresses, with nodes representing objects whose states (position, velocity, and appearance) evolve over time. The DGNN component refines the node associations across frames, ensuring that the objects are consistently tracked, even under complex conditions such as occlusions or rapid movements. This integration of detection and tracking into a single pipeline provides a unified solution that adapts to dynamic environments, ensuring reliable and real-time performance. The framework also incorporates XAI techniques such as Grad-CAM, Grad-CAM++, and Eigen-CAM to enhance the system's transparency. These methods visualize the key regions of the image that influence the model's detection decisions, making the system more interpretable and allowing users to gain insights into the critical factors driving object identification and tracking.

% \clearpage

% Adding pseudocode:

\begin{center}
% \newpage
\scalebox{1}{
\begin{algorithm}[H]
\footnotesize
\SetAlgoLined
\KwIn{Video frames $F = \{f_1, f_2, ..., f_n\}$, Pre-trained YOLO11 model weights, DGNN model, ROI settings}

\KwOut{Annotated frames with tracked small objects and FPS}

% Initialization
\textbf{Initialize} YOLO11 model $M_{YOLO}$ with pre-trained weights\;

\textbf{Initialize} Dynamic Graph Neural Network (DGNN) model $M_{DGNN}$\;

\textbf{Load} video file or camera stream\;

\textbf{Initialize} graph $G$ for tracking\;

\textbf{Define} region of interest (ROI)\;

\While{frames available in video stream}{
    \textbf{Capture} next frame $f_t$\;

    \textbf{Define ROI} for object detection and tracking\;

    \textbf{Run YOLO11} on $f_t$ to get detections $\{d_1, d_2, ..., d_m\}$\;

    \If{detections exist}{
        \textbf{Extract node features} (center coordinates, width, height) for each detection\;

        \ForEach{node $n_i$ in detections}{
            \textbf{Refine bounding box} to constrain dimensions\;

            \textbf{Update position and velocity} for $n_i$\;
        }

        \textbf{Update graph} $G$\;

        \ForEach{node pair $(n_i, n_j)$ in graph}{
            \If{distance $< \text{threshold}$ or velocity difference $< \text{threshold}$}{
                \textbf{Add edge} $(n_i, n_j)$\;
            }

            \Else{
                \textbf{Remove edge} $(n_i, n_j)$\;
            }
        }

        \textbf{Construct edge index tensor} from $G$\;

        \textbf{Refine associations} using $M_{DGNN}$\;

        \textbf{Annotate frame} with slim bounding boxes, object centers, and FPS value\;
    }

    \Else{
        \textbf{Log} ``No detections in this frame"\;
    }

    \textbf{Display annotated frame} to user\;

    \If{'q' key pressed}{
        \textbf{Break loop}\;
    }
}

\textbf{Release resources} (video stream, memory)\;

\caption{DGNN-YOLO framework for small object detection and tracking}
\label{algo:1}
\end{algorithm}
}
\end{center}

\subsubsection{DGNN-YOLO Workflow}
As described in Algorithm~\ref{algo:1}, the DGNN-YOLO framework processes video frames using the YOLOv11 detection module to identify objects and generate bounding boxes. The detected objects are then represented as nodes in a dynamic graph. The edges in the graph model the relationships between objects, capturing spatial-temporal features such as proximity, velocity, and appearance similarity. The dynamic nature of the graph allows it to adapt as objects move, enter, or exit the frame, with new edges being added or removed based on changes in these relationships.

The graph is then processed by the DGNN, which refines the associations between objects by updating the node embeddings. This dynamic update ensures that the model effectively tracks objects across frames, even when they are occluded or exhibit unpredictable movement. The refined graph, incorporating spatial and temporal dependencies, is the foundation for accurate object tracking.

\subsubsection{Advantages of Integration}
The DGNN-YOLO framework addresses the limitations of traditional detection and tracking methods, which typically rely on separate, disjointed stages for each task. By integrating detection and tracking into a unified framework, DGNN-YOLO offers several key advantages. The YOLOv11 model ensures accurate and reliable object detection even under challenging conditions, while the DGNN dynamically updates the relationships between objects over time. This allows the framework to handle environmental changes, such as occlusions or variations in object motion, more effectively.

The adaptability of the DGNN allows for real-time updates to the graph, reflecting changes in object positions, velocities, and appearances across frames. This integration enhances detection accuracy and tracking reliability, enabling the system to perform consistently in complex, high-density environments like traffic surveillance~\citep{wang2019fast}. DGNN-YOLO avoids the inefficiencies and limitations of traditional multi-stage pipelines and provides more robust performance in dynamic real-world conditions by using a single, unified pipeline.

\subsection{YOLO11-Based Detection Mechanism}
Figure~\ref{fig:whole_process} also shows the YOLOv11 architecture within the DGNN-YOLO framework, which is optimized for accurate small-object detection. YOLOv11 comprises three main components: the backbone, the neck, and the head.

The backbone of YOLOv11 consists of a series of stacked convolutional layers that extract multi-scale features from the input image. These features capture the spatial hierarchies of the objects, allowing the model to detect fine-grained details even in cluttered environments. The neck processes these extracted features and uses advanced modules like Spatial Pyramid Pooling Fast (SPPF) and Channel-to-Parallel Spatial Attention (C2PSA) to enhance the feature representations. SPPF aggregates features from different receptive fields to capture global and local contexts, while C2PSA sharpens spatial and positional awareness, improving the model's ability to detect small objects in dense environments.

\subsubsection{Backbone: Feature Extraction and Multi-Scale Representation}
The backbone of YOLOv11 consists of a series of stacked convolutional layers that extract multi-scale features from the input image. These layers are designed to capture spatial hierarchies and fine-grained details, enabling the model to detect small objects even in dense and cluttered environments. YOLOv11 uses advanced convolutional operations to detect fine-scale patterns, which are crucial for accurately localizing and classifying small objects such as pedestrians, cyclists, and motorbikes, often seen in traffic surveillance footage.

YOLOv11’s backbone goes beyond traditional convolutions by incorporating dilated convolutions and strided convolutions, which expand the receptive field while maintaining computational efficiency. This design helps the network capture both local and global context, ensuring that objects at varying scales are detected without sacrificing accuracy. Additionally, multi-scale feature fusion is employed to combine features from different scales, which is especially beneficial for handling both large and small objects within the same scene.

\subsubsection{Neck: Feature Refinement with Advanced Modules}
The neck of YOLOv11 processes the feature maps extracted by the backbone, further refining them for object detection. Key modules in the neck include Spatial Pyramid Pooling Fast (SPPF) and Channel-to-Parallel Spatial Attention (C2PSA), which are designed to enhance the model’s ability to detect small and occluded objects in dynamic and complex environments like traffic scenes.

\begin{enumerate}
    \item \textbf{Spatial Pyramid Pooling Fast (SPPF)}: SPPF aggregates features from multiple receptive fields, allowing the model to capture both global context (large objects or features) and local context (small objects or fine details). This multi-scale pooling helps YOLOv11 handle objects of varying sizes effectively, improving its robustness in traffic environments with mixed object scales.
    
    \item \textbf{Channel-to-Parallel Spatial Attention (C2PSA)}: C2PSA sharpens the spatial awareness of the network by employing a parallel attention mechanism that selectively focuses on important regions of the image. This allows the model to dynamically adjust its attention to areas more likely to contain objects, especially small and occluded ones. By improving spatial and positional precision, C2PSA enhances YOLOv11’s ability to detect and localize objects accurately in densely packed environments, ensuring small objects are correctly identified even when partially obscured.
\end{enumerate}

\subsubsection{Head: Object Detection and Localization}
The head of YOLOv11 generates the final detection outputs using the refined feature maps from the neck. The head predicts the bounding boxes, confidence scores, and class labels for each detected object.

\begin{enumerate}
    \item \textbf{Bounding Boxes}: Defined by the center coordinates \( (x, y) \) and dimensions \( (w, h) \), these parameters represent the predicted location and size of each object. YOLOv11’s head uses these coordinates to accurately localize objects, even in cases of occlusion or when objects are located close together.
    
    \item \textbf{Confidence Scores}: These scores reflect the model’s confidence in the presence of an object within a predicted bounding box. The confidence score \( C_i \) quantifies the likelihood that the box contains an object and is used to assess the detection quality.
    
    \item \textbf{Class Labels}: The class label \( L_i \) identifies the object type, such as “car,” “pedestrian,” or “motorbike.” YOLOv11’s head uses the softmax activation function to classify the object based on the feature embeddings generated by the backbone and neck.
\end{enumerate}

To ensure precision, Non-Maximum Suppression (NMS) is applied. NMS removes redundant and overlapping bounding boxes by selecting the one with the highest confidence score and discarding others with lower scores. This is crucial in dense traffic environments where objects are often clustered closely together, ensuring that only the most accurate and distinct predictions are retained.

\subsubsection{Detection Outputs}  
The outputs generated by YOLO11 consisted of \textit{bounding boxes}, \textit{confidence scores}, and \textit{class labels}. Bounding boxes are defined by coordinates \((x, y, w, h)\), where \(x\) and \(y\) denote the center of the box, and \(w\) and \(h\) specify its width and height, respectively. Confidence scores indicate the likelihood of an object being present within a bounding box, while class labels identify the object type, such as ``car," ``pedestrian," or ``motorbike." To enhance precision, non-maximum suppression (NMS) was applied to remove redundant and overlapping detections, retaining only the most confident predictions for each object. This refinement step is particularly crucial in high-density scenarios with frequent overlapping detection. The filtered outputs, including refined bounding boxes, confidence scores, and class labels, form the basis for constructing a dynamic graph, enabling robust spatial-temporal modeling in the subsequent tracking stage.

\subsection{Dynamic Graph Construction}
Figure~\ref{fig:whole_process} shows how the DGNN-YOLO framework, the dynamic graph construction, is a fundamental process that enables effective object tracking across frames. This graph is updated continuously in real-time to reflect the changing state of the scene, such as the appearance and movement of objects. Each detected object in the frame is represented as a node in the graph, and edges capture the relationships between objects. The graph is updated at each frame, ensuring that the system tracks objects consistently, even as they move, occlude each other, or change their position and interaction in the scene.

The dynamic graph structure allows the system to adapt to the ongoing changes in the traffic scenario. New nodes are created as new objects enter the frame, and as objects leave or become irrelevant, their corresponding nodes are removed. Simultaneously, the relationships between objects, modeled as edges, are recalculated to reflect the updated spatial and temporal interactions. This updating mechanism ensures that the graph accurately represents the real-time scene, facilitating robust tracking and the prediction of object behaviors.

\subsubsection{Node Representation}
Each node \( n_i \) in the graph represents a detected object and contains a feature vector that integrates multiple attributes to describe the object’s properties. These attributes include spatial, motion, and appearance features, each contributing to the node’s ability to represent the object’s characteristics and interactions in the scene.

\begin{enumerate}
    \item \textbf{Spatial Features}: The spatial features are crucial for accurately locating the object within the image. The center represents the coordinates \( (x_i, y_i) \) of the bounding box as well as its dimensions \( (w_i, h_i) \), which correspond to the width and height of the bounding box. These features help localize the object’s position in the scene, allowing the model to detect and track objects accurately, even when partially occluded or in crowded areas.
    \item \textbf{Motion Features}: The motion features capture the object’s movement across frames. These are represented by the velocity components \( v_{x_i} \) and \( v_{y_i} \), which describe how the object moves in the x and y directions. By calculating the difference in position over consecutive frames, these features model the object’s trajectory and movement pattern, providing valuable temporal information for tracking.
    \item \textbf{Appearance Features}: The appearance of each object is encoded through feature embeddings derived from the YOLOv11 detection model. These embeddings represent the visual characteristics of the object, such as its texture, color, and shape. These features are essential for distinguishing between objects with similar spatial locations but different visual appearances, helping to resolve ambiguities, especially in scenarios with occlusions or overlapping objects.
\end{enumerate}

By combining these spatial, motion, and appearance features, each node effectively captures both static and dynamic attributes of the object, making it possible to model the object’s state and its interactions with other objects in the scene. The representation is flexible, enabling the system to adapt to both visual and motion cues across frames.

\subsubsection{Edge Construction}
In the dynamic graph, the edges represent the interactions between nodes and model the spatial-temporal relationships between objects. These edges are dynamically updated to capture how objects interact over time based on three key factors: proximity, velocity similarity, and appearance similarity.

\begin{enumerate}
    \item \textbf{Proximity}: Proximity is measured by the Euclidean distance between the centers of two objects’ bounding boxes. The distance \( d_{ij} = \| (x_i, y_i) - (x_j, y_j) \|_2 \) captures how spatially close two objects are in the scene. Objects close to each other are more likely to interact, and proximity is a strong indicator of potential collisions or groupings, which is especially important in traffic surveillance where vehicles, pedestrians, and cyclists often overlap or move in close formation.
    
    \item \textbf{Velocity Similarity}: The velocity similarity between two objects is calculated using the velocity vectors \( v_i(t) \) and \( v_j(t) \) of objects \( i \) and \( j \) at time \( t \). The difference in velocity \( \Delta v_{ij} = \| v_i(t) - v_j(t) \|_2 \) quantifies how similarly the two objects are moving. Objects with similar velocities are more likely to follow each other or maintain consistent relative motion, which helps predict their trajectories and ensures that the graph structure reflects their evolving relationships over time.
    
    \item \textbf{Appearance Similarity}: Appearance similarity is based on the feature embeddings \( f_i \) and \( f_j \) extracted from the YOLOv11 model, which describe the visual characteristics of the objects. The cosine similarity between these feature vectors, \( \text{CosineSim}(f_i, f_j) \), is computed to assess how visually similar two objects are, even when they are at different positions or have different velocities. This is especially useful in scenarios where objects are occluded or appear in similar visual contexts, allowing the system to correctly associate objects despite their changing positions.
\end{enumerate}

By combining these three factors—proximity, velocity similarity, and appearance similarity—the edges in the graph dynamically update to reflect the strength and nature of interactions between objects. The adjacency matrix \( A_t \), which encodes the edge connections between nodes, is recalculated in each frame based on these relationships. This allows the graph to adapt to changes in object interactions, ensuring that the system accurately tracks objects and maintains consistent associations even as their positions and motions evolve over time.

\subsubsection{Graph Updates}
The dynamic graph is updated at each frame to ensure that it accurately reflects the scene's current state. As new objects are detected, new nodes are added to the graph, and when objects leave the scene or become occluded, their corresponding nodes are removed. This update process is essential for keeping the graph efficient and relevant to the current scene.

Edge recalculation is another critical part of graph updates. As objects move or interact, the edges between nodes are recalculated based on the updated spatial, motion, and appearance features. This ensures that the relationships between objects are always accurate and up-to-date, which is crucial for maintaining the integrity of the tracking process. Furthermore, the propagation component of the graph ensures that information is consistently spread across the nodes. When objects interact or influence each other, their state is updated, and this influence is propagated throughout the graph. This helps maintain continuity in the tracking process, even when objects are occluded or overlapping. The propagation mechanism allows the system to track objects across frames, even as new objects enter or existing ones exit, ensuring that the system adapts to changes in the scene in real-time.

The system can track objects accurately and consistently over time by updating the graph dynamically and continuously propagating information, even in highly complex environments where occlusions, rapid motion, and dense object distributions are common. This real-time adaptability makes the DGNN-YOLO framework highly effective for traffic surveillance, where traffic scenarios are constantly changing, and real-time object tracking is essential.

\subsection{DGNN-Based Tracking}
The DGNN refines object tracking by leveraging spatial-temporal dependencies within the graph. It uses graph convolutional layers to model objects' interactions, capturing their spatial relationships and temporal dynamics. The DGNN dynamically updates the node and edge representations by processing the graph at each frame, ensuring the tracking remains consistent, even in occlusions, rapid motion, or crowded environments. This approach allows DGNN-YOLO to effectively handle complex interactions, such as objects entering or exiting a scene or overlapping trajectories, making it robust against the challenges of real-world traffic scenarios.

\subsubsection{Graph Convolution Layers}
The DGNN employs graph convolutional layers to aggregate information from neighboring nodes, allowing the network to model spatial-temporal relationships effectively. Each layer updates the node features using the adjacency matrix \(A_t\), which encodes the edge relationships between nodes. The forward propagation at layer \(l\) is defined as:
\begin{equation}
H^{(l+1)} = \sigma(A_t H^{(l)} W^{(l)})
\end{equation}
where \(H^{(l)}\) represents the node features at layer \(l\), 
\(W^{(l)}\) denotes the learnable weight matrix for layer \(l\), and \(\sigma\) denotes a non-linear activation function, such as ReLU.

This iterative process propagates information across the graph and refines node embeddings in each layer. By leveraging the adjacency matrix, the network captures relationships such as proximity, velocity similarity, and appearance similarity between objects. These refined embeddings enable the DGNN to distinguish between objects with similar appearances, thereby enhancing the robustness of object tracking in crowded or visually ambiguous environments~\citep{wang2019fast}.

\subsubsection{Output Predictions}
The DGNN outputs refined node embeddings and updated edge weights, which assign unique tracking IDs to objects. These IDs ensure consistent identification of objects across consecutive frames. DGNN effectively handles challenges such as occlusions, abrupt object movements, and overlapping trajectories by modeling evolving spatiotemporal relationships. This capability enables the system to maintain robust tracking even in complex traffic environments.

\subsection{Loss Functions}
DGNN-YOLO adopts a combined loss function to optimize both detection and tracking tasks. The detection loss (\(L_{\text{det}}\)) ensures accurate object localization and classification, while the tracking loss (\(L_{\text{track}}\)) minimizes temporal inconsistencies in object trajectories, enhancing smoothness and reliability across frames. Inspired by approaches like FairMOT, which integrate detection and re-identification into a unified framework, DGNN-YOLO balances these objectives using a weighted combination of the two losses~\citep{zhang2021fairmot}. The total loss function is defined as:
\begin{equation}
L = \lambda_{\text{det}} L_{\text{det}} + \lambda_{\text{track}} L_{\text{track}}
\end{equation}
where \(\lambda_{\text{det}}\) and \(\lambda_{\text{track}}\) are weighting factors that control the relative contributions of detection and tracking losses. This formulation enables DGNN-YOLO to perform robustly in challenging scenarios, such as dense traffic and highly dynamic environments.

\subsection{XAI Techniques}
To improve the interpretability of the DGNN-YOLO model and provide insights into its decision-making process, we employed three XAI techniques: Grad-CAM, Grad-CAM++, and Eigen-CAM. These methods highlight regions of interest in the input image, allowing users to visualize which features contribute most to the predictions of DGNN-YOLO.

\subsubsection{Grad-CAM}
Gradient-weighted class activation mapping (Grad-CAM) leverages the gradients of the target class flowing into the final convolutional layer to generate a localization map. This map identifies regions most relevant to the model's prediction, making it an essential tool for understanding spatial attention. Mathematically, the localization map \( L_{\text{Grad-CAM}}^c \) for a given class \( c \) is computed as:
\begin{equation}
L_{\text{Grad-CAM}}^c = \text{ReLU} \left( \sum_k \alpha_k^c A^k \right) 
\end{equation}
where \( A^k \) denotes the \( k \)-th feature map, and \( \alpha_k^c \) is the weight representing the importance of the feature map \( A^k \) to class \( c \). The weight \( \alpha_k^c \) is calculated as:
\begin{equation}
\alpha_k^c = \frac{1}{Z} \sum_{i,j} \frac{\partial y^c}{\partial A_{i,j}^k}
\end{equation}
Here, \( y^c \) is the output score for class \( c \), \( Z \) normalizes the spatial dimensions, and \( i, j \) index the spatial locations. The ReLU function ensures the map highlights only positive contributions corresponding to the regions that increase the target class score.

\subsubsection{Grad-CAM++}
Grad-CAM++, an extension of Grad-CAM, addresses its limitations, particularly in scenarios involving multiple instances of the same class within an image. By refining the computation of weights \( \alpha_k^c \), Grad-CAM++ assigns higher importance to regions that strongly influence the output while better handling overlapping or small objects. The revised weight calculation in Grad-CAM++ is given by:
\begin{equation}
\alpha_k^c = \frac{\sum_{i,j} \frac{\partial^2 y^c}{\partial (A_{i,j}^k)^2}}{2 \sum_{i,j} \frac{\partial^2 y^c}{\partial (A_{i,j}^k)^2} + \sum_{i,j} \frac{\partial y^c}{\partial A_{i,j}^k}} 
\end{equation}
This enhancement improves the resolution and sensitivity of the localization map, making it more effective in identifying fine-grained details critical for small-object detection.

\subsubsection{Eigen-CAM}
Eigen-CAM introduces an unsupervised approach to interpretability by analyzing the principal components of the activation maps. Unlike Grad-CAM and Grad-CAM++, which rely on backpropagation, Eigen-CAM focuses on the intrinsic properties of feature maps to highlight dominant regions contributing to the model’s decisions. The localization map \( L_{\text{Eigen-CAM}} \) is derived as:
\begin{equation}
L_{\text{Eigen-CAM}} = \text{EigenVec}(A) 
\end{equation}
where \( \text{EigenVec}(A) \) represents the principal eigenvector of the activation matrix \( A \). This technique identifies significant variations within the feature maps, offering a holistic view of the model’s focus without requiring class-specific gradients.

\section{Experiments and Results}
\label{sec:experiments}
This section comprehensively evaluates the proposed DGNN-YOLO framework for small object detection and tracking.

\subsection{Dataset}
The experiments were conducted on the \textit{i2 Object Detection Dataset}, specifically curated for traffic surveillance tasks. This dataset includes diverse traffic scenes under various conditions, such as occlusions, varying lighting, and dense object distributions. The dataset comprises 50,000 labeled images with 24 object classes, including vehicles (\textit{Car, Bus, Truck}), non-motorized objects (\textit{Cycle, Rickshaw}), and pedestrians (\textit{person}). The data split was organized into two main sets. The training set comprised 80\%, consisting of 40,000 images, whereas the validation set comprised the remaining 20\%, totaling 10,000 images.

\begin{figure}[!ht]
    \centering
    \includegraphics[width=0.8\linewidth]{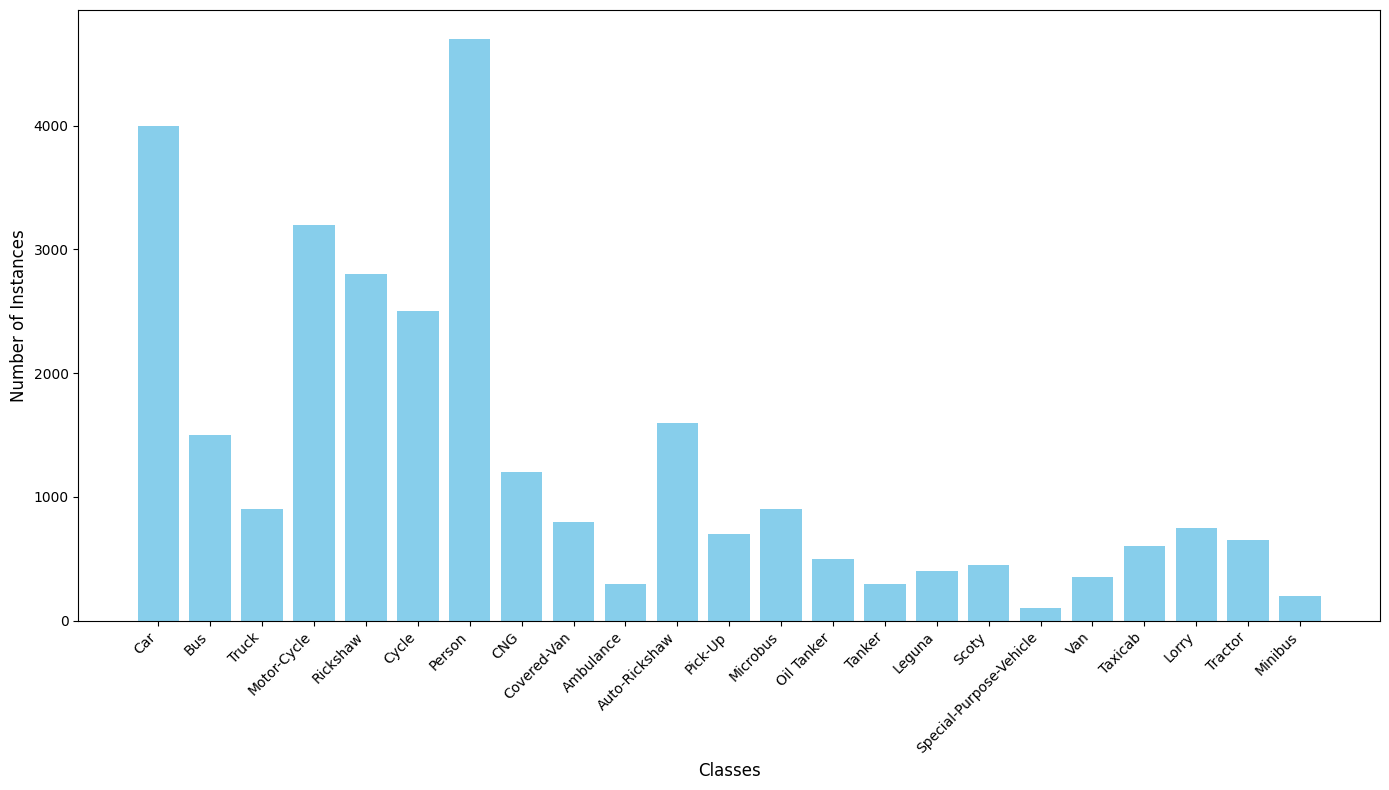}
    \caption{Class-wise distribution of the \textit{i2 Object Detection Dataset}.}
    \label{fig:class_distribution}
\end{figure}

Data augmentation techniques such as flipping, scaling, and brightness adjustments have been applied to address class imbalance. The class distribution, illustrated in Figure~\ref{fig:class_distribution}, highlights the predominance of certain classes, such as \textit{Car} and \textit{Bus}, while others, such as \textit{ambulance}, are underrepresented. This distribution emphasizes the importance of robust training strategies for effectively handling imbalanced datasets.

\subsection{Experimental Setup}
The DGNN-YOLO framework was implemented in Python using the PyTorch library and trained on an NVIDIA RTX 4050 GPU. The training consisted of 40 epochs, with an input image resolution of $280 \times 280$ pixels. The YOLO11 module was initialized with pre-trained COCO weights to leverage the learned features for object detection, whereas the DGNN module was trained from scratch. This setup ensures that the DGNN effectively captures the dynamic relationships between objects, improving the tracking performance in challenging traffic scenarios. Training parameters, such as learning rate and batch size, were dynamically adjusted during the experiment to optimize the model convergence and performance.

\subsection{Evaluation Metrics}  
The performance of the DGNN-YOLO framework was assessed using a comprehensive set of metrics to evaluate its accuracy and robustness for small-object detection and tracking.

\subsubsection{Precision}  
Precision measures the proportion of correctly predicted positive cases out of all the predicted positives. This reflects the ability of the system to minimize false positives. Mathematically, the precision is expressed as:  
\begin{equation}
\text{Precision} = \frac{\text{TP}}{\text{TP} + \text{FP}}
\end{equation}  
where TP represents true positives (i.e., correctly detected objects), and FP denotes false positives (i.e., incorrectly detected objects).

\subsubsection{Recall}  
Recall captures the ratio of true positives to all actual positive cases, highlighting the system's ability to detect relevant objects. It is defined as:  
\begin{equation}
\text{Recall} = \frac{\text{TP}}{\text{TP} + \text{FN}}
\end{equation}  
where FN refers to false negatives, that is, objects missed by the detection system.

\subsubsection{Mean Average Precision (mAP)}  
Mean Average Precision evaluates the trade-off between precision and recall across different confidence thresholds. It is computed as the mean of the overall average precision (AP) of the object classes. AP is calculated as the area under the precision-recall curve:  
\begin{equation}
\text{AP} = \int_{0}^{1} \text{Precision}(\text{Recall}) \, d(\text{Recall}).
\end{equation}  
Two specific metrics are used: \textbf{mAP@0.5}, which evaluates AP at a fixed Intersection over Union (IoU) threshold of 0.5, and \textbf{mAP@0.5:0.95}, which averages AP over multiple IoU thresholds ranging from 0.5 to 0.95 in increments of 0.05. The latter provides a holistic assessment of the detection performance of a system.

\subsubsection{Interpretability Metrics}
We utilized the following evaluation metrics to assess the interpretability of the XAI techniques applied to DGNN-YOLO.

\textbf{Faithfulness:}
Faithfulness measures how well the explanations align with the model's decision-making process. An explanation is considered faithful if removing or altering the most important regions identified by the XAI method significantly impacts the model's predictions. This metric evaluates the causal relationship between the highlighted regions and the model's outputs, ensuring the reliability of the provided explanations.

\textbf{Flipping:}
Flipping quantifies the robustness of the model's explanations. It is calculated as the proportion of cases where altering the regions identified as important (e.g., masking or modifying pixel intensities) leads to a change in the model's prediction. A higher flipping rate indicates that the identified regions are crucial for the model's decision, thus validating the explanation's relevance.

\textbf{Complexity:}
Complexity assesses the cognitive load required to understand the explanations generated by XAI methods. It considers factors such as the number of regions or features highlighted in an explanation and their spatial distribution. Simpler and more intuitive explanations are desirable, as they improve the comprehensibility and usability of the model in real-world applications.

\textbf{Comprehension of 80\% of Attributes (\%):}
This metric measures the extent to which the explanations cover 80\% of the critical attributes necessary for the model's prediction. It is reported as a percentage, reflecting how effectively the explanations capture the key decision-making factors while minimizing irrelevant or redundant information. Higher values indicate better alignment with the underlying attributes influencing the model's outputs.

\subsection{Results and Analysis}
This section analyzes the performance of the DGNN-YOLO framework using quantitative metrics and qualitative observations. The results demonstrate the capability of DGNN-YOLO to detect and track small objects under various challenging conditions effectively.

\subsubsection{Comparative Experiments}
We compared the proposed DGNN-YOLO (YOLO11+DGNN) model with various object detection models, including standard YOLO versions and Faster R-CNN. As shown in Table~\ref{tab:benchmark_results}, YOLO11 with DGNN achieves the highest precision (0.8382) and recall (0.6875) among all models. Additionally, it demonstrated superior performance for mAP@0.5 and mAP@0.5:0.95, with values of 0.7830 and 0.6476, respectively. These results validate the effectiveness of incorporating the DGNN into YOLO11, particularly in improving the detection accuracy for small objects. 

The performance of YOLO11 without DGNN was lower, with a precision of 0.8176 and a recall of 0.5248. The mAP@0.5 and mAP@0.5:0.95 for YOLO11 were also significantly lower than YOLO11 with DGNN, indicating the critical role of DGNN in improving spatial relationship modeling. Other models, such as YOLO10, YOLO9, and YOLO8, exhibit progressively lower performance metrics, whereas Faster R-CNN achieves competitive results but still lags behind YOLO11+DGNN. This analysis highlights the ability of DGNN-YOLO to outperform existing models in addressing the challenges of small-object detection and tracking.

\begin{table}[!ht]
\footnotesize
\centering
\caption{Benchmarking results for DGNN-YOLO against YOLO variants and Faster R-CNN. \textbf{Bold} indicates the best performance.}
\label{tab:benchmark_results}
\begin{tabular}{lcccccc}
\hline
\textbf{Model}          & \textbf{Precision} & \textbf{Recall} & \textbf{mAP@0.5} & \textbf{mAP@0.5:0.95} & \textbf{Training time (s)} & \textbf{Parameters (in millions)} \\ \hline
Faster R-CNN            & 0.7582             & 0.6075          & 0.7830           & 0.4976                & 40342.34          & $\approx 18.9$    \\ 
YOLO5                   & 0.7789             & 0.5191          & 0.5917           & 0.4679                & 26894.90          & $\approx 9.2$     \\ 
YOLO8                   & 0.7055             & 0.6085          & 0.6175           & 0.4926                & 21967.85          & $\approx 11.2$    \\ 
YOLO9                   & 0.8354             & 0.5330          & 0.6152           & 0.4903                & 19504.32          & $\approx 7.3$     \\ 
YOLO10                  & 0.7786             & 0.5096          & 0.6045           & 0.4852                & 21967.85          & $\approx 8.1$     \\ 
YOLO11                  & 0.8176             & 0.5248          & 0.6107           & 0.4871                & 17040.80          & $\approx 9.5$     \\ 
\textbf{Proposed DGNN-YOLO}      & \textbf{0.8382}            & \textbf{0.6875}          & \textbf{0.7830}           & \textbf{0.6476}                & 118344.00      & \textbf{$\approx 20.1$}  \\ \hline
\end{tabular}
\end{table}
\normalsize

\begin{figure}[!ht]
    \centering
    \includegraphics[width=\linewidth]{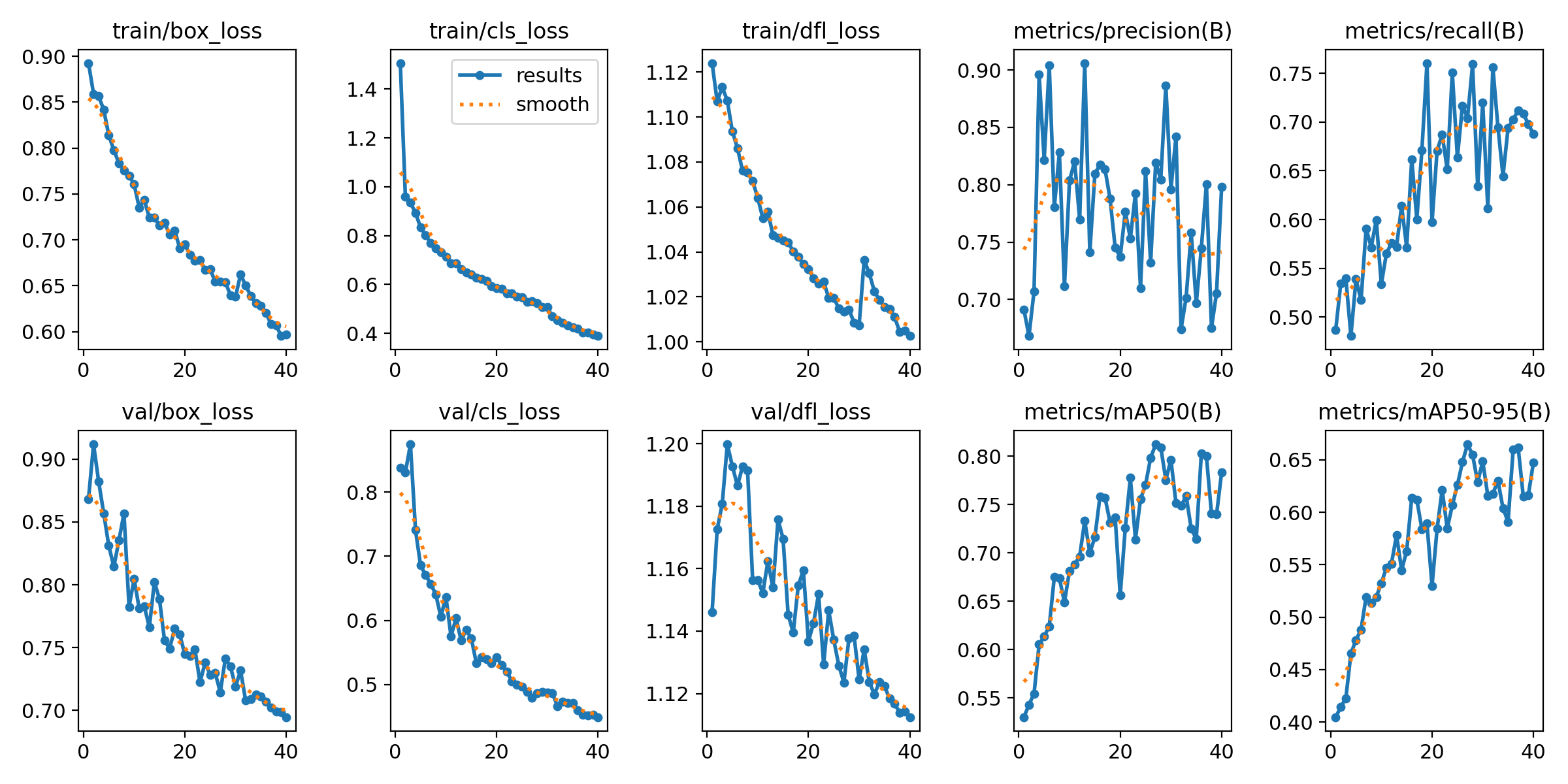}
    \caption{Performance metrics across epochs for the proposed DGNN-YOLO.}
    \label{fig:training_metrics}
\end{figure}

Figure~\ref{fig:training_metrics} presents the training and validation performances of the DGNN-YOLO model over 40 epochs, showing its optimization and detection capabilities. The top row displays the training losses for bounding box regression, classification, and distributional focal loss, whereas the bottom row shows the corresponding validation losses. All the losses exhibited a steady decline, indicating effective learning and convergence. The precision and recall metrics showed stable improvements with steadily increasing recall, reflecting a better identification of TP. The mAP@0.5 and mAP@0.5:0.95 also improve consistently, with mAP@0.5 reaching around 0.8 and mAP@0.5:0.95 nearing 0.65 by the final epoch, suggesting strong detection performance. These trends confirm the ability of the DGNN-YOLO model to learn effectively and generalize across datasets.

\begin{figure}[!ht]
    \centering
    \includegraphics[width=0.6\linewidth]{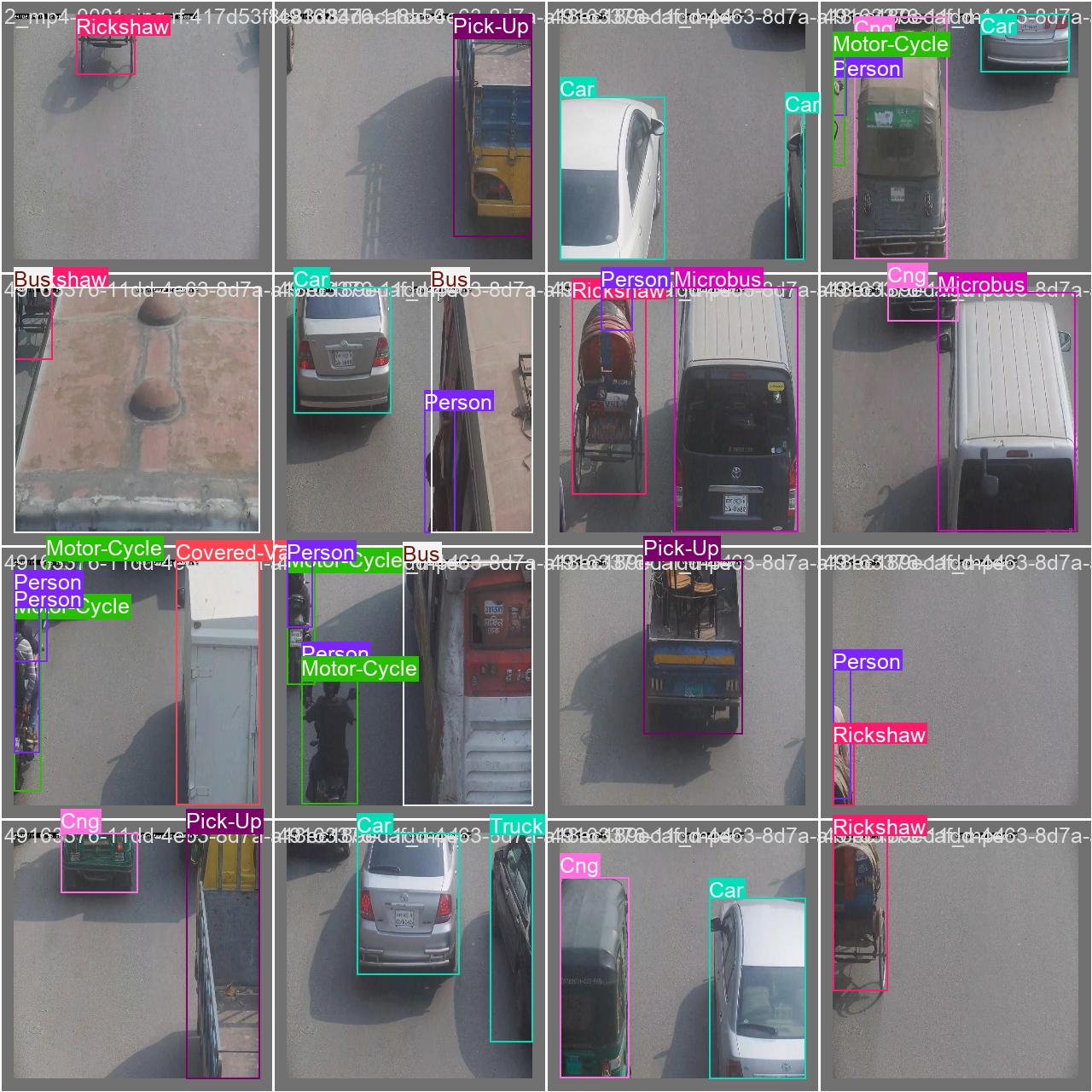}
    \caption{Validation results of DGNN-YOLO showing small object detection and tracking.}
    \label{fig:validation_predictions}
\end{figure}

\subsubsection{Validation Analysis}
Figure~\ref{fig:validation_predictions} demonstrates the detection capabilities of the DGNN-YOLO model, demonstrating its performance in identifying and localizing multiple objects across different categories in a real-world traffic scene. The detected objects, such as ``Rickshaw," ``Car," ``Bus," ``Motor-Cycle," ``Person," and others, are enclosed within bounding boxes with class labels. The model effectively identifies overlapping objects and differentiates between small and large instances, thereby highlighting its robustness in handling complex scenarios with diverse object types, sizes, and spatial arrangements. Accurate detection indicates the model’s ability to generalize effectively and its suitability for multi-object detection tasks in real-world environments.
\begin{figure}[!ht]
    \centering
    \includegraphics[width=0.8\linewidth]{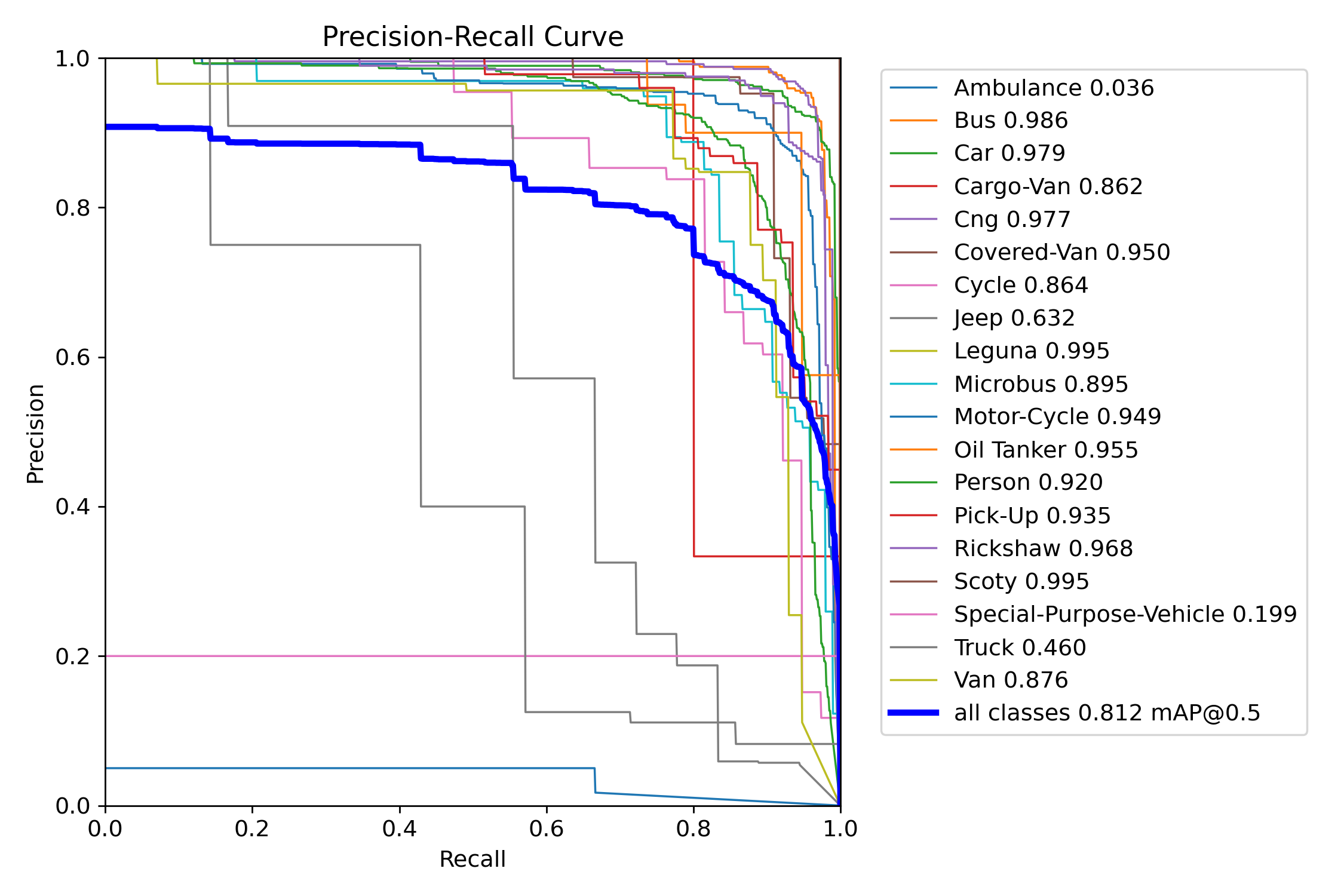}
    \caption{Precision-recall curve of DGNN-YOLO for class-wise evaluation.}
    \label{fig:precision_recall}
\end{figure}

\subsubsection{Precision-Recall Analysis}
Figure~\ref{fig:precision_recall} illustrates the performance of the DGNN-YOLO model across multiple object classes. Notably, the model achieved an average mAP@0.5 of 0.812, highlighting its robust detection capabilities across most categories. High-performing classes such as ``Leguna" (0.995), ``Scooty" (0.995), and ``Bus" (0.986) demonstrate the model's proficiency in recognizing distinct object types, potentially benefiting from pronounced class-specific features. However, there is notable underperformance in classes like ``Ambulance" (0.036) and ``Special-Purpose-Vehicle" (0.199), which may be attributed to limited training samples or insufficient feature differentiation. The steep drop in precision at higher recall levels for some classes, such as ``Truck" (0.460), indicates challenges in maintaining consistent confidence levels under exhaustive detections. DGNN-YOLO demonstrated strong generalization, although optimization for underrepresented or visually similar classes could further enhance its real-world applicability.

\begin{figure}[!ht]
\centering
\includegraphics[width=1.1\linewidth]{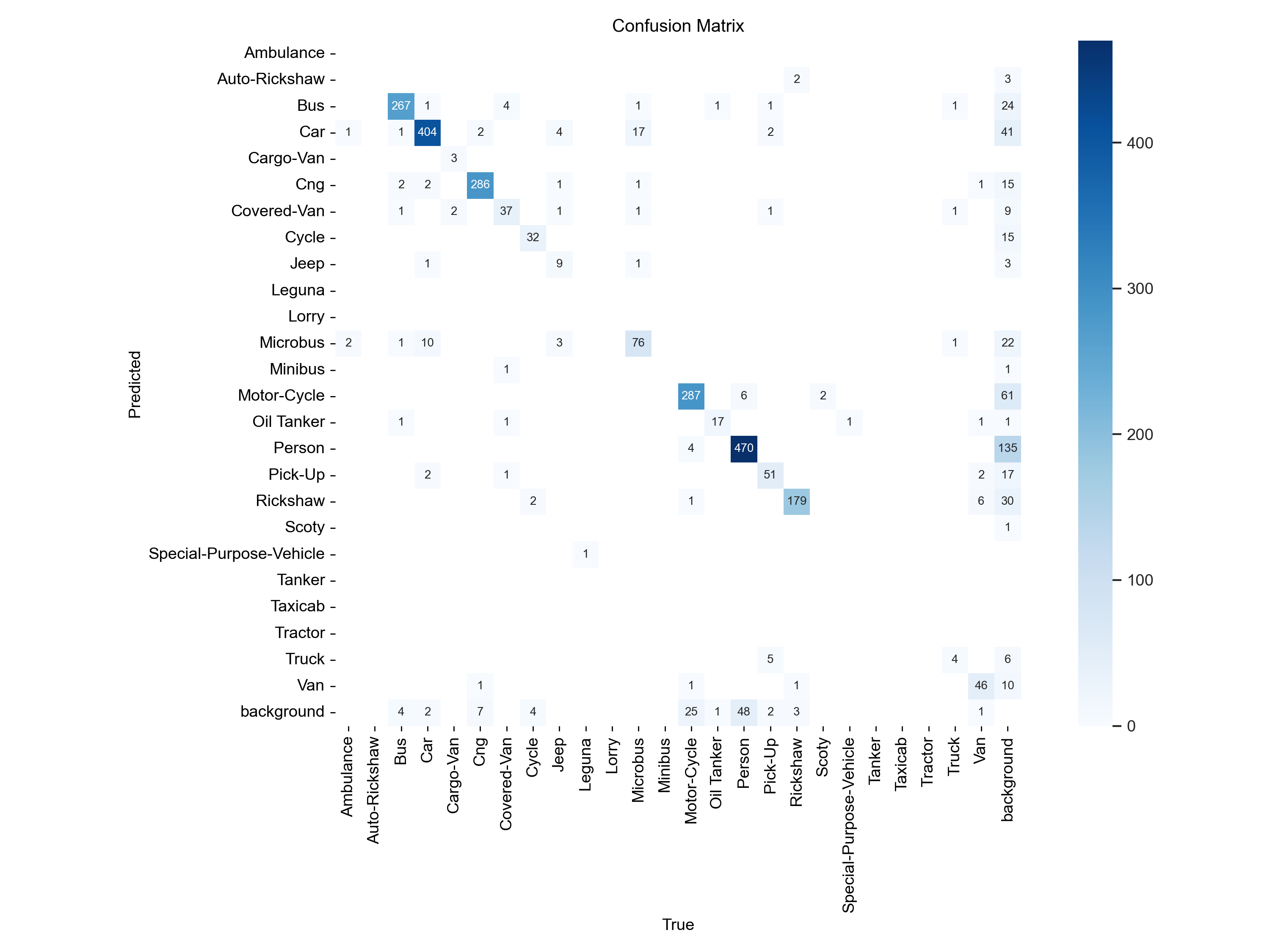}
\caption{Class-wise confusion matrix for DGNN-YOLO framework.}
\label{fig:confusion_matrix}
\end{figure}

Figure~\ref{fig:confusion_matrix} presents the confusion matrix for the DGNN-YOLO model, illustrating its classification performance across multiple classes. Each row corresponds to the predicted labels, and each column represents the true label. The diagonal entries indicate the correctly classified instances for each class, with darker shades of blue signifying higher counts, reflecting better performance. The off-diagonal entries represent misclassifications, with lighter shades of blue highlighting lower error frequencies. For example, the model performs well in identifying classes such as ``Person," ``Bus," and ``Motor-Cycle," evidenced by the high values on the diagonal for these categories. However, it struggles with certain smaller classes like ``Jeep" and ``Special-Purpose-Vehicle," which have lower diagonal values and noticeable misclassification counts. The accompanying color bar visually represents the density of predictions, with the highest counts reaching above 400 for certain classes like ``Car" and ``Person." This matrix underscores both the strengths and weaknesses of DGNN-YOLO, demonstrating its effectiveness in recognizing frequently occurring categories while hinting at potential areas of improvement for less-represented or visually similar classes.

\begin{table}[!ht]
\centering
\caption{Ablation study results for different configurations of the DGNN framework.}
\begin{tabular}{lcccc}
\hline
\textbf{Configuration}             & \textbf{mAP} & \textbf{Precision} & \textbf{Recall} & \textbf{Frames per second} \\ \hline
\textbf{Full DGNN-YOLO Framework}                & \textbf{0.716}        & \textbf{0.776}              & \textbf{0.636}           & \textbf{$\approx 60$} \\ 
DGNN w/o Appearance Embeddings      & 0.690        & 0.760              & 0.610           & $\approx 62$ \\ 
DGNN w/o Velocity Similarity        & 0.700        & 0.765              & 0.620           & $\approx 63$ \\ 
DGNN w/o Temporal Features          & 0.680        & 0.740              & 0.600           & $\approx 65$ \\ 
Constant Edge Weights              & 0.705        & 0.770              & 0.625           & $\approx 64$ \\ \hline
\end{tabular}
\label{tab:ablation_study}
\end{table}

\subsubsection{Ablation Studies}
The ablation studies offer a comprehensive analysis of the proposed YOLO11-DGNN framework, evaluating its robustness under various configurations and highlighting the contributions of its spatial and temporal components. Table~\ref{tab:ablation_study} presents the quantitative metrics (mAP, precision, recall, and frames per second) across five configurations: full DGNN framework, DGNN without appearance embeddings, DGNN without velocity similarity, DGNN without temporal features, and constant edge weights. The full DGNN framework achieved the highest mAP (0.716), precision (0.776), and recall (0.636), with an efficient processing rate of $\approx60$ FPS, confirming its superior performance. Removing appearance embeddings or velocity similarity resulted in moderate declines in mAP and recall, underscoring the importance of these factors in maintaining robust object associations. Excluding temporal features has a more pronounced impact, leading to the lowest mAP (0.680) and recall (0.600), demonstrating the critical role of temporal dynamics in tracking moving objects. Using constant edge weights reduces the adaptive capability of the framework, causing slight drops in precision and recall.

\begin{table}[!ht]
\centering
\caption{Error Metrics Across Configurations of the DGNN Framework.}
\begin{tabular}{lccc}
\hline
\textbf{Configuration}             & \textbf{MAE} & \textbf{RMSE} & \textbf{MAPE (\%)} \\ \hline
\textbf{DGNN-YOLO}                & \textbf{10.5}        & \textbf{23.0}              & \textbf{15.0}           \\ 
DGNN w/o Spatial            & 11.0        & 22.5              & 16.0           \\ 
DGNN w/o Temporal           & 12.0        & 25.0              & 17.5           \\ \hline
\end{tabular}
\label{tab:error_metrics}
\end{table}

Table~\ref{tab:error_metrics} highlights the critical impact of spatial and temporal components on the performance of the DGNN-YOLO framework. The YOLO11-DGNN configuration achieved the best results, with the lowest MAE (10.5), RMSE (23.0), and MAPE (15.0\%), demonstrating the importance of integrating both components. Removing spatial features (DGNN w/o Spatial) slightly increases MAE and MAPE, indicating a mild performance decline owing to reduced object differentiation in cluttered scenes. The absence of temporal features (DGNN w/o Temporal) results in the highest errors across all metrics, as temporal dynamics are essential for maintaining object identities across frames, particularly under occlusions or overlapping scenarios. These findings emphasize the necessity of both features for achieving robust detection and tracking in complex environments.

\begin{figure}[!ht]
    \centering
    \includegraphics[width=0.6\linewidth]{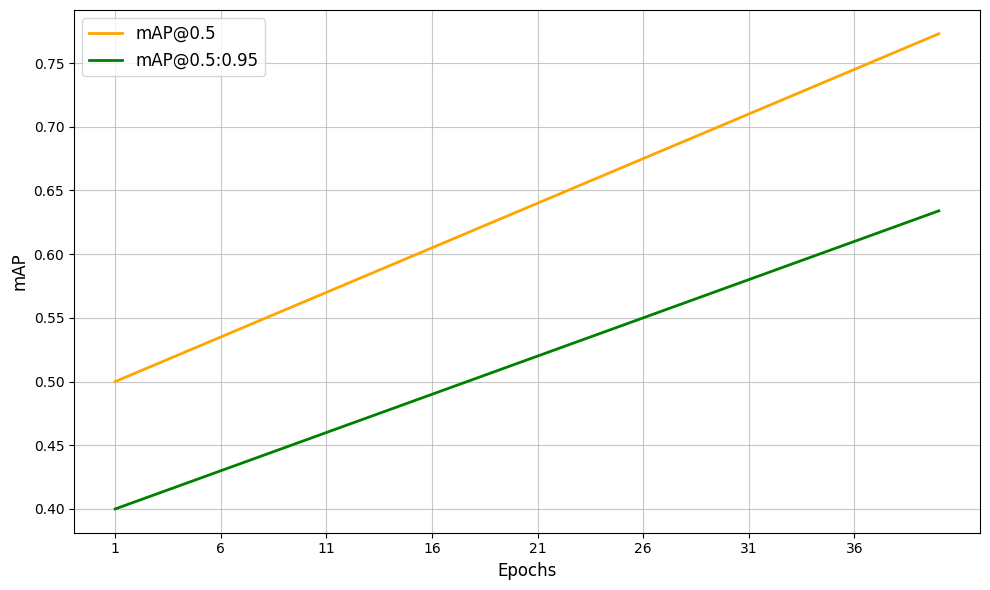}
    \caption{Qualitative results of the DGNN-YOLO model showing real-world detection scenarios.}
    \label{fig:mapepochs}
\end{figure}

\begin{figure}[!ht]
    \centering
    \includegraphics[width=0.6\linewidth]{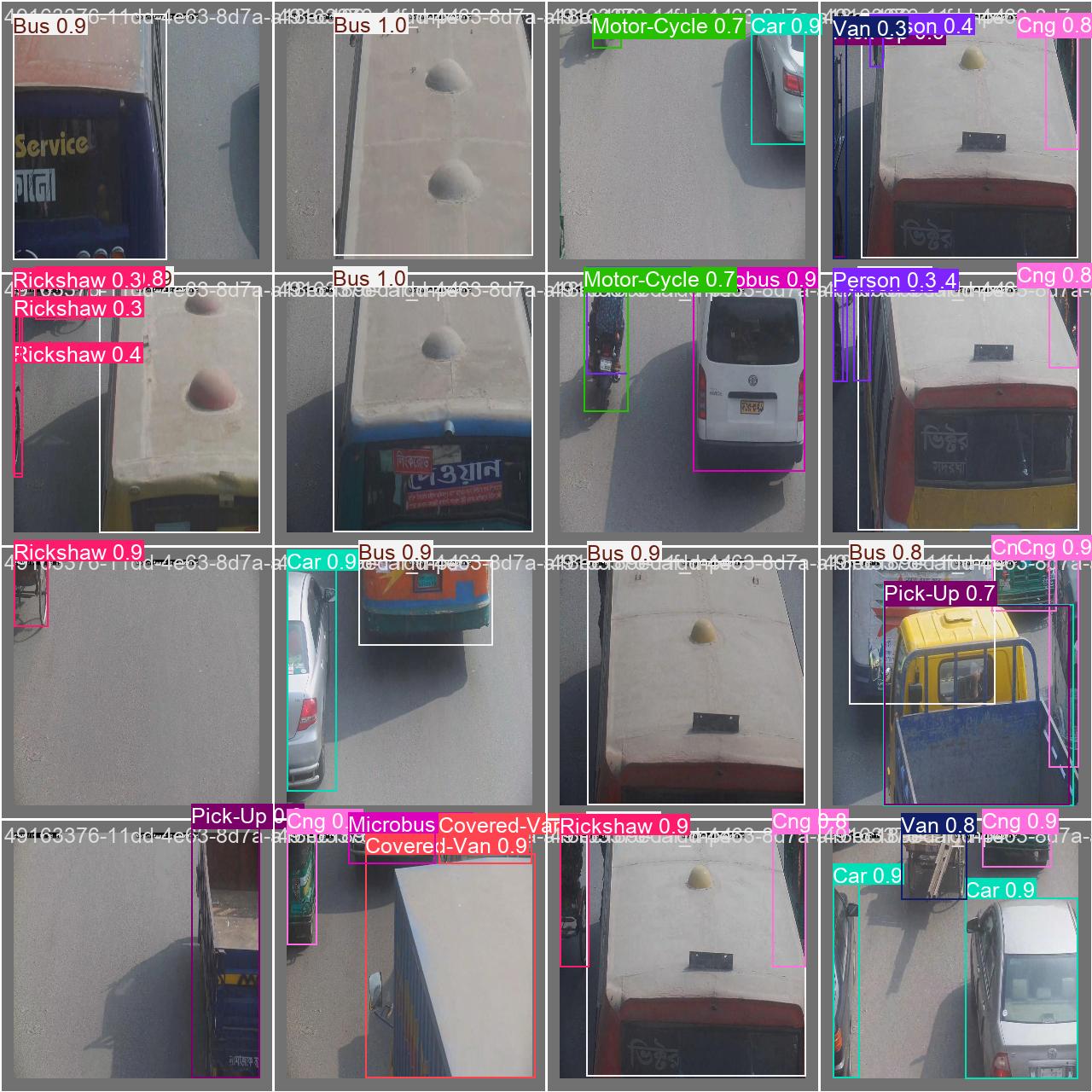}
    \caption{Qualitative results of the DGNN-YOLO model showing real-world detection scenarios.}
    \label{fig:qualitative_results}
\end{figure}

\subsubsection{mAP Improvements over Epochs}
Figure~\ref{fig:mapepochs} shows the trends in mAP@0.5 and mAP@0.5:0.95 across 40 training epochs. The steady and consistent improvement observed in both metrics reflects the effectiveness of the learning process and the ability of the DGNN-YOLO model to generalize. Specifically, the \textbf{mAP@0.5} exhibited significant growth, reaching a maximum of 0.716 by the final epoch, demonstrating robust detection performance at a fixed IoU threshold of 0.5. In contrast, \textbf{mAP@0.5:0.95} shows gradual but steady improvements, achieving a value of 0.575 by the end of the training, which highlights the capability of the DGNN-YOLO model to maintain consistent performance across varying IoU thresholds, further reinforcing its small object detection and tracking ability. Overall, the consistent upward trend in these metrics validates the convergence of the model during training and demonstrates its proficiency in fine-tuning detection and tracking capabilities across complex traffic scenarios.

\subsubsection{Interpretability Analysis}
\begin{figure}[!ht]
    \centering
    \includegraphics[width=0.8\linewidth]{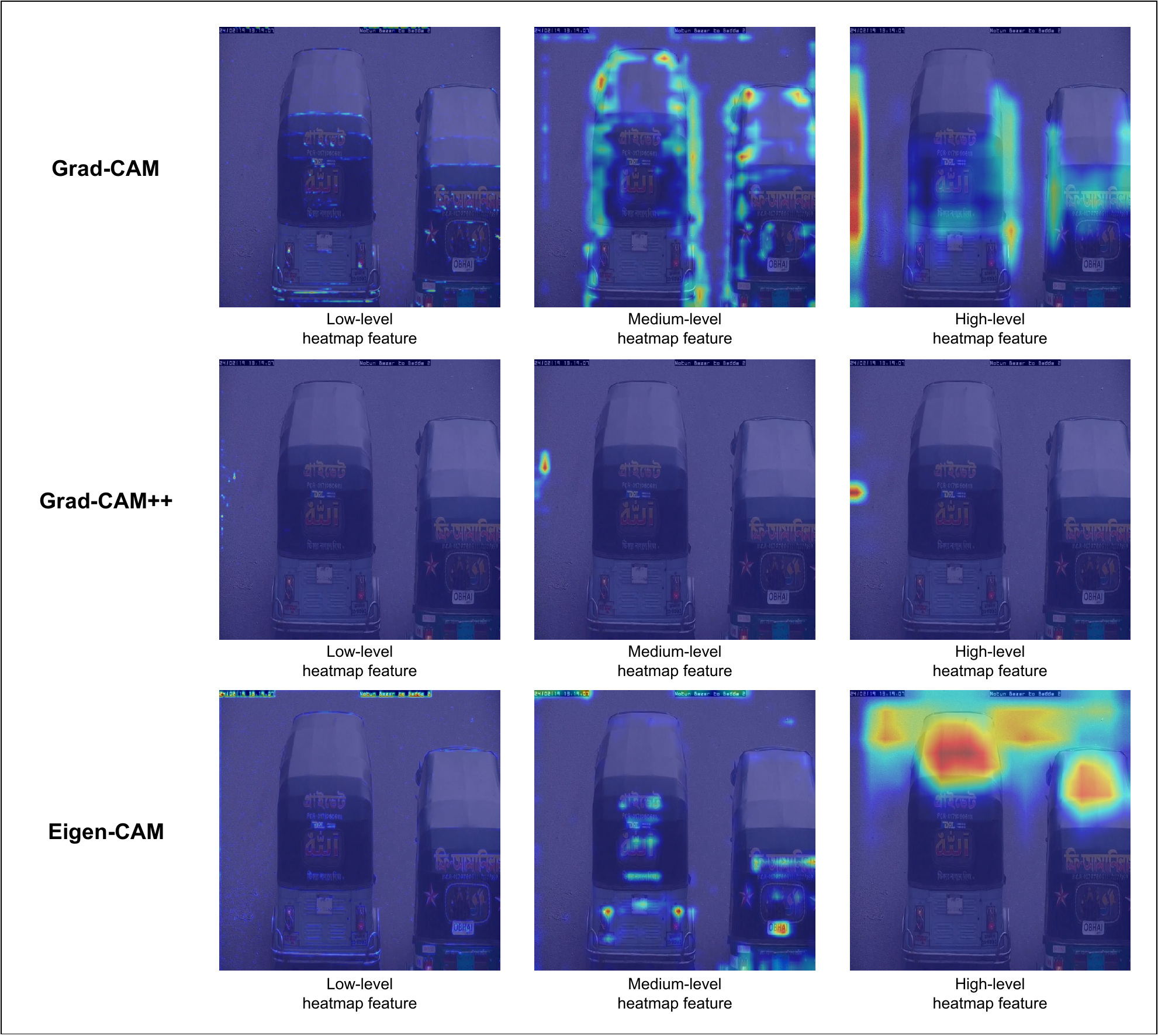}
    \caption{Comparison of Grad-CAM, Grad-CAM++, and Eigen-CAM applied on proposed DGNN-YOLO.}
    \label{fig:xai_result}
\end{figure}

Figure~\ref{fig:xai_result} visually compares Grad-CAM, Grad-CAM++, and Eigen-CAM, illustrating each algorithm's focus regions when interpreting small object detection and tracking in traffic surveillance. Grad-CAM provides a broader view of the areas DGNN-YOLO attends to, showing a more general understanding of the object locations. Grad-CAM++ refines this attention, focusing on more specific areas of interest within the objects, thereby improving localization. Eigen-CAM, on the other hand, further enhances the precision by concentrating on the most relevant features of the object, offering the highest level of detail. This progression in attention refinement highlights the strengths of each method in terms of its interpretability and ability to provide meaningful insights into the model’s decision-making process.

\begin{table}[h!t]
\centering
\caption{Comparison of XAI Algorithms based on Faithfulness and Complexity Metrics.}
\begin{tabular}{lcccc}
\hline
\textbf{XAI Algorithm} & \textbf{Faithfulness} & \textbf{Flipping} & \textbf{Complexity} & \textbf{Comprehension of 80\% of attributes (\%)} \\ \hline
Grad-CAM    & 2.446046  & 2.929214  & 1.326923  & 39.378779  \\ 
Grad-CAM++  & 1.644412  & 2.024183  & 1.326013  & 38.382997  \\ 
Eigen-CAM   & 1.563907  & 2.655633  & 1.293600  & 39.281843  \\ \hline
\end{tabular}
\label{tab:xai_comparison}
\end{table}

Table~\ref{tab:xai_comparison} quantitatively assesses the XAI algorithms using four key metrics. Grad-CAM achieves the highest Faithfulness score of 2.446, reflecting its strong alignment with predictions of DGNN-YOLO, while Grad-CAM++ and Eigen-CAM score 1.644 and 1.563, respectively. The flipping metric, which evaluates consistency under input perturbations, highlights Grad-CAM’s performance with 2.929, followed by Eigen-CAM at 2.655 and Grad-CAM++ at 2.024. For the complexity metric, a lower score signifies simpler and more comprehensible explanations; Eigen-CAM achieves the lowest score of 1.293, indicating its advantage in interpretability, while Grad-CAM and Grad-CAM++ have scores of 1.326 and 1.326, respectively. Finally, the comprehension metric, which measures the percentage of attributes needed to explain 80\% predictions of DGNN-YOLO, is nearly consistent across all methods, with Grad-CAM at 39.38\%, Eigen-CAM at 39.28\%, and Grad-CAM++ at 38.38\%.

\subsubsection{Qualitative Evaluation}
Figure~\ref{fig:qualitative_results} showcases qualitative results obtained using the DGNN-YOLO framework, highlighting its robustness in real-world traffic scenarios. DGNN-YOLO detects and tracks small, overlapping objects, such as motorcycles and rickshaws, even in dense traffic conditions. Key observations include accurately detecting small and large objects with well-defined bounding boxes and class labels and effectively handling occlusions and overlapping objects, particularly among vehicles closely positioned in traffic. Additionally, DGNN-YOLO demonstrates robust performance under suboptimal lighting conditions, showcasing its adaptability in challenging environments. These results illustrate the practical applicability of DGNN-YOLO in real-time traffic surveillance, where accurate detection and tracking are critical for monitoring and decision-making.

%% main text

%%%%%%%%%%%%%%%%%%%%%%%%%%%%%%%%%%%%%%%%%%
\section{Conclusions}
\label{sec:conclusion}
The DGNN-YOLO framework brings together YOLO11 and DGNNs to solve the important problem of detecting and tracking small, occluded objects in urban traffic. Our experiments show strong results, with a precision of 0.8382, a recall of 0.6875, and mAP@0.5:0.95 of 0.6476, outperforming existing methods in challenging traffic situations. The main strength of DGNN-YOLO is its ability to adapt to changing traffic conditions through real-time graph updates, allowing it to track objects reliably, even when they are occluded or moving erratically. Additionally, tools like Grad-CAM, Grad-CAM++, and Eigen-CAM make the model more interpretable, which helps build trust for practical use in the real world.

Despite its strong performance, DGNN-YOLO has some limitations. While it works well in dense traffic, it struggles in extreme weather conditions like heavy rain or fog because it relies only on visual data. Due to dataset imbalances, the model also has difficulty distinguishing visually similar or underrepresented classes, such as ambulances. To improve this, future work will focus on integrating sensors like LiDAR to help the model perform better in low-visibility situations. Another important goal is to deploy DGNN-YOLO on edge devices, allowing for real-time processing in environments with limited resources, essential for scaling smart city solutions.

Looking ahead, there are several key areas for improvement. We plan to test DGNN-YOLO on various datasets, including MOTChallenge and KITTI, to make it more adaptable to different traffic scenarios. Additionally, we aim to optimize the model for edge deployment by using techniques like quantization and pruning to make it more efficient. We will also explore semi-supervised learning to improve the detection of rare classes, reducing our dependence on manually labeled data. By addressing these challenges, DGNN-YOLO will not only enhance urban mobility but also contribute to the development of intelligent, scalable transportation systems.

%%%%%%%%%%%%%%%%%%%%%%%%%%%%%%%%%%%%%%%%%%

%% The Appendices part is started with the command %\appendix;
%% appendix sections are then done as normal sections
%\appendix
\section*{Data Availability Statement}
The dataset utilized in this study, the \textit{i2 Object Detection Dataset}, is accessible at the following link: \\
\href{https://universe.roboflow.com/data-48lkx/i2-waurd}{https://universe.roboflow.com/data-48lkx/i2-waurd}.

\section*{Funding Sources}
This research did not receive any specific grants from funding agencies in the public, commercial, or not-for-profit sectors.

\section*{CRediT Authorship Contribution Statement}
\textbf{Shahriar Soudeep:} Conceptualization, Investigation, Methodology, Formal Analysis, Writing - Original Draft.
\textbf{Md Abrar Jahin:}  Conceptualization, Methodology, Formal Analysis, Writing - Original Draft, Validation.
\textbf{M. F. Mridha:} Validation, Supervision.

\section*{Declaration of Competing Interest}{The authors declare that they have no known competing financial interests or personal relationships that could have appeared to influence the work.}

\section*{Acknowledgement}{The authors would like to thank the Advanced Machine Intelligence Research Lab - AMIR Lab for supervision and resources.}
%% If you have bibdatabase file and want bibtex to generate the
%% bibitems, please use
%%

%\bibliographystyle{elsarticle-num} 

% \bibliographystyle{apalike}
\bibliographystyle{plainnat}
\bibliography{main}
% \bibliography{Exported Items}

%% else use the following coding to input the bibitems directly in the
%% TeX file.

% \begin{thebibliography}{00}

% %% \bibitem[Author(year)]{label}
% %% Text of bibliographic item

% \bibitem[ ()]{}

% \end{thebibliography}
\end{document}